\documentclass[letterpaper, 10 pt, conference]{ieeeconf}  

\usepackage{amsmath,amsfonts}
\usepackage{algorithmic}
\usepackage{array}
\usepackage[caption=false,font=normalsize,labelfont=sf,textfont=sf]{subfig}
\usepackage{textcomp}
\usepackage{stfloats}
\usepackage{pifont}
\usepackage{cite}
\makeatletter
\let\NAT@parse\undefined
\makeatother
\usepackage{lineno,hyperref}
\usepackage{booktabs} 
\usepackage{amsmath,array,booktabs,tabularx}
\hypersetup{
	colorlinks=true, 
	linkcolor=blue,  
	citecolor=blue, 
	urlcolor=red     
}
\usepackage{url}
\usepackage{verbatim}
\usepackage{graphicx}
\usepackage{xcolor}
\def\BibTeX{{\rm B\kern-.05em{\sc i\kern-.025em b}\kern-.08em
		T\kern-.1667em\lower.7ex\hbox{E}\kern-.125emX}}
\usepackage{balance}

\IEEEoverridecommandlockouts                              

\title{\LARGE \bf
Closed-Form Cartesian Forward Kinetostatics for Spatial Multi-Segment Tendon-Driven Continuum Robots}

\author{Ke Wu$^{1,3}$, Fangju Yang$^{1}$, Xiaohui Zhang$^{2}$, Junda He$^{3}$, Guanjun Bao$^{3}$, Jingang Yi$^{4}$, Jian S. Dai$^{5}$
}

\begin{document}

\maketitle
\thispagestyle{empty}
\pagestyle{empty}

\begin{abstract}
Forward kinetostatics of spatial tendon-driven continuum robots
typically requires a nonlinear equilibrium solve for each actuation
input. This paper develops a force-to-Cartesian-configuration model
with a closed-form solution in quadratures for spatial multi-segment
robots under tendon actuation. The Cartesian backbone centerline and
accumulated material twist serve as generalized coordinates, from
which the strain measures and tendon geometry are derived.
Variational equilibrium yields explicit axial and bending relations
and establishes zero equilibrium material twist within the proposed
model for admissible longitudinal non-helical routing. The solution
is propagated segment by segment without an iterative equilibrium
solve, while retaining axial deformation, spatially varying axial
and bending stiffnesses and tendon-routing diameter, and
segment-dependent tendon participation. Numerical comparisons with a
full-strain geometric variable-strain model (GVS) yield maximum
length-normalized tip-position discrepancies of
\(8.91\times10^{-6}\) and \(1.01\times10^{-5}\) for the single- and
three-segment robots, respectively. Mean evaluation times of \(1.52\,\mathrm{\mu s}\) and
\(2.94\,\mathrm{\mu s}\), with corresponding speedups of
approximately \(1864\times\) and \(3348\times\) over the baseline,
demonstrate the computational advantage of the explicit
force-to-configuration mapping in the reported benchmark.
\end{abstract}



\section{Introduction}

Forward kinetostatics is fundamental to the design
\cite{rao2021model}, workspace characterization
\cite{rucker2011statics}, motion planning
\cite{wang2024quasistatic}, and model-based control
\cite{della2023model,cao2017workspace} of tendon-driven continuum robots (TDCRs). It
determines the equilibrium robot configuration produced by tendon
actuation \cite{rucker2011statics}, thereby establishing the mapping
from actuation space to task space. Unlike rigid-link robots, TDCRs
deform continuously along their backbones
\cite{gilbert2021mathematical}, and their equilibrium configurations
depend on distributed mechanical and geometric properties
\cite{rao2021model}. The resulting actuation-to-configuration mapping is
therefore a coupled kinematic and static problem. A broad range of
forward kinetostatic formulations have consequently been developed,
differing fundamentally in how they represent the continuously
deformable backbone \cite{armanini2023soft}.

Existing formulations differ in both the deformation assumptions
imposed on the backbone and the variables used to describe its
configuration. Piecewise-constant-curvature (PCC) models describe
each section using a small set of arc parameters
\cite{webster2010design} and recover the spatial robot pose through
successive geometric transformations \cite{jones2006kinematics}.
Beyond PCC, mechanics-based formulations use rotation, strain, or
Cartesian position fields as their primary variables. Planar large-deflection models use cross-sectional rotation fields as
their primary variables \cite{gravagne2003large}. Cosserat rod models describe the backbone through
coupled position, orientation, and strain fields
\cite{rucker2011statics}, whereas geometric variable-strain models
approximate distributed strains using a finite set of basis
coefficients \cite{renda2020geometric}. Cartesian formulations choose
global position or displacement fields as their primary unknowns and
derive strain measures from their spatial derivatives. Representatives include spline models based on Cartesian control points
\cite{luo2020spline}, finite-element models based on nodal
displacements \cite{bieze2018finite}, and absolute-state models based
on Cartesian nodal positions augmented by orientation variables
\cite{sadati2019reduced}. Cartesian variables thereby express the
backbone geometry directly in the coordinates used for task-space
planning, sensing, and control \cite{bieze2018finite}, whereas strain
fields enter rod constitutive laws directly and intrinsically describe
local deformation \cite{simo1991geometrically}. These choices relocate rather than eliminate geometric nonlinearity:
Cartesian formulations inherit nonlinear strain--configuration
relations, whereas strain-based formulations require nonlinear spatial
reconstruction of the robot pose
\cite{gilbert2021mathematical}.

Accordingly, most general mechanics-based formulations recover the
equilibrium configuration numerically. Continuous Cosserat models solve
nonlinear spatial boundary-value problems
\cite{rucker2011statics}. Geometric variable-strain models reduce the
distributed mechanics to nonlinear algebraic equations in strain
coordinates \cite{renda2020geometric}, whereas Cartesian
finite-element models solve constrained systems of nodal variables
\cite{bieze2018finite}. The computational cost and convergence of these
methods depend on the spatial resolution, numerical scheme, and initial
estimate \cite{rao2021model}. Specialized Cosserat implementations can
achieve real-time performance \cite{till2019realtime}. Nevertheless,
the actuation-to-configuration relation remains implicit and generally
requires a new equilibrium solve for each actuation input
\cite{rao2021model}. PCC models reduce the problem dimension through a
small number of section variables \cite{webster2010design}. When axial
stretch or curvature varies within a section because of nonuniform
stiffness or tendon-routing geometry, additional section subdivision is
required \cite{armanini2023soft}. Existing formulations therefore balance computational compactness
against the resolution of distributed deformation.

Closed-form forward force-to-shape models remain confined to
restrictive settings. Kato et al.~\cite{kato2014tendon} map proximal
tendon tensions to lumped-cell curvatures and assemble the resulting
multi-section posture. Roesthuis and
Misra~\cite{roesthuis2016steering} derive explicit unloaded
tension-to-curvature relations under the PCC assumption, but require a
nonlinear solve in the general loaded case. Related closed-form small-deflection beam solutions have been derived
for initially straight robots with prescribed tendon displacements
\cite{oliver2019continuum}. More
recently, Wu et al.~\cite{wu2026lightweight} derive a closed-form
planar force-to-shape model accommodating nonuniform stiffness,
variable tendon routing, and axial deformation, while treating gravity
semi-analytically through Adomian decomposition \cite{duan2012review}. To the best of our knowledge, its extension to spatial deformation still remains open. 

This work develops an analytical kinetostatic framework mapping
tendon forces directly to Cartesian configurations of spatial
multi-segment TDCRs.
Using the Cartesian centerline and accumulated material twist
as generalized coordinates, the formulation retains axial
deformation, spatially varying axial and bending stiffnesses,
variable tendon-routing diameter, and segment-dependent
tendon participation. The main contributions are threefold.
First, under the stated loading and routing conditions,
variational equilibrium yields explicit axial and bending
relations and establishes zero material twist within the model.
Second, a closed-form solution in quadratures reconstructs
the continuous Cartesian backbone segment by segment without
an iterative equilibrium solve.
Third, comparisons with geometric
variable-strain (GVS) \cite{renda2020geometric} demonstrate close
agreement in backbone configurations and distributed strains
for single- and three-segment robots at substantially lower
evaluation costs.
Across the sampled tendon-force inputs, both robots exhibit
maximum length-normalized tip-position discrepancies below
\(1.02\times10^{-5}\), average evaluation times below
\(0.003\,\mathrm{ms}\), and speedups of up to approximately
\(3348\times\) over GVS.
These results support repeated forward-model evaluations
in motion planning and model-based control.

\section{Problem Statement}
\begin{figure*}[!t]
    \centering \includegraphics[width=1.0\textwidth, trim=0cm 6cm 0cm 4cm,]{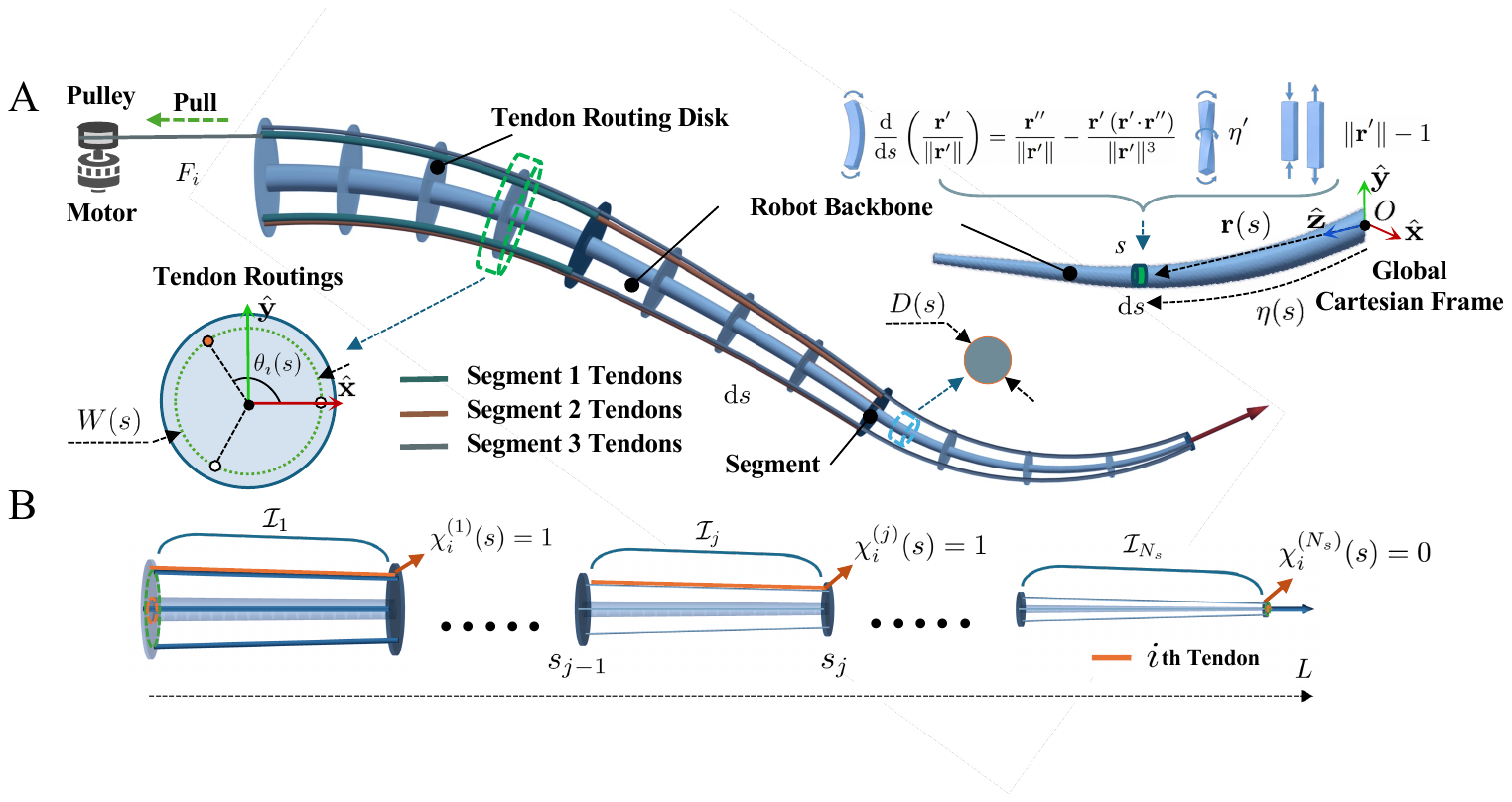}
    \caption{The studied tendon-driven continuum robot.}
    \label{fig:studied-manipulator}
\end{figure*}

\subsection{The studied manipulator}
As shown in Fig.~\ref{fig:studied-manipulator}, we consider a multi-segment tendon-driven continuum robot with a straight and untwisted stress-free reference configuration. Let \(s\in[0,L]\) denote the reference arc-length coordinate of the robot backbone in 3D space. Multiple tendons are routed longitudinally around the backbone, with a segment-dependent subset routed through each segment. The robot has spatially varying axial, isotropic bending, and torsional
stiffnesses \(EA(s)\), \(EI(s)\), and \(GJ(s)\), respectively,
together with a spatially varying tendon-routing diameter \(W(s)\).
Here, \(A(s)\), \(I(s)\), and \(J(s)\) denote the cross-sectional area,
second moment of area, and torsional constant.

\subsection{Objective}

The objective is to derive a closed-form force-to-shape formulation
that maps the applied tendon forces directly to the equilibrium
three-dimensional Cartesian configuration of the robot while
accounting for segment-wise tendon routing and spatially varying
mechanical and geometric properties. 

\subsection{Modeling Conditions}
\label{MC}

The model is developed under the following conditions:
\begin{itemize}
    \item The backbone is modeled as a linearly elastic slender rod
    made of an isotropic material.

    \item Tendons are routed longitudinally and non-helically.

    \item Tendon friction is neglected.

    \item External loads other than tendon actuation are neglected.
\end{itemize}

\section{Methodology}

\subsection{Generalized Coordinates and Tendon Geometry}
\label{subsec:generalized-coordinates}
Let
\(\mathcal{F}=\{O;\hat{\mathbf{x}},\hat{\mathbf{y}},\hat{\mathbf{z}}\}\)
be a fixed right-handed global Cartesian frame, with \(O\) at the robot base,
\(\hat{\mathbf{z}}\) aligned with the prescribed base tangent, and
\(\hat{\mathbf{x}}\) and \(\hat{\mathbf{y}}\) spanning the base cross-sectional
plane and defining its zero angular reference. \(\mathbf r:[0,L]\rightarrow\mathbb R^3\) denotes the backbone centerline
expressed in \(\mathcal F\). The accumulated material twist \(\eta(s)\) is measured relative to a rotation-minimizing frame along the centerline. 
The Cartesian generalized coordinates and their clamped-base
conditions are
\begin{equation}
\small
\begin{aligned}
\mathbf q(s)
=
\begin{bmatrix}
\mathbf r^{\mathbf T}(s)&\eta(s)
\end{bmatrix}^{\mathbf T},\ 
\mathbf r(0)
=\mathbf 0,\ 
\frac{\mathbf r'(0)}
{\|\mathbf r'(0)\|}
=\hat{\mathbf z},\ 
\eta(0)=0.
\end{aligned}
\label{eq:cartesian-generalized-coordinate}
\end{equation}
A prime denotes differentiation with respect to \(s\).
Motivated by the slender backbone geometry, transverse shear
is neglected in the present formulation~\cite{grassmann2022fas},
while axial deformation, bending, and material twist are retained. For a regular centerline satisfying \(\|\mathbf r'(s)\|>0\),
the cross-sectional normal consequently coincides with
\(\mathbf r'/\|\mathbf r'\|\). The axial, bending, and torsional
strain measures are therefore obtained directly from \(\mathbf q\):
\begin{equation*}
\small
\begin{aligned}
\text{axial strain:}\quad
&\|\mathbf r'\|-1,\ 
\text{torsional strain:}\quad
\eta',\\
\text{bending strain:}\quad
&
\frac{\mathrm d}{\mathrm ds}
\left(
\frac{\mathbf r'}{\|\mathbf r'\|}
\right)
=
\frac{\mathbf r''}{\|\mathbf r'\|}
-
\frac{
\mathbf r'
\left(
\mathbf r'\!\cdot\!\mathbf r''
\right)
}{
\|\mathbf r'\|^3
}.
\end{aligned}
\end{equation*}
Let \(0=s_0<s_1<\cdots<s_{N_s}=L\) partition the backbone into
\(N_s\) segments, and let
\(\Gamma_j\subseteq\{1,\ldots,n\}\) contain the indices of the tendons
routed through segment \(j\). Define
\begin{equation*}
\small
\begin{aligned}
\mathcal I_j
&=
\begin{cases}
[s_{j-1},s_j), & j=1,\ldots,N_s-1,\\
[s_{N_s-1},L], & j=N_s,
\end{cases}\\
\chi_i(s)
&=
\begin{cases}
1, & s\in\mathcal I_j,\ i\in\Gamma_j
     \text{ for some }j,\\
0, & \text{otherwise}.
\end{cases}
\end{aligned}
\end{equation*}
The generalized coordinates are piecewise smooth, with
\(\mathbf r\), \(\mathbf r'/\|\mathbf r'\|\), and \(\eta\) continuous
across segment interfaces. The mechanical properties may be piecewise
smooth, whereas \(W(s)\) is continuous and piecewise \(C^1\), with
\(W'\in L^\infty(0,L)\). For longitudinal non-helical routing, let \(\theta_i\) denote the
constant material angular position of tendon \(i\), measured from
\(\hat{\mathbf x}\) in the base cross-sectional plane. Its radial
direction is reconstructed from the generalized coordinates through
\begin{equation}
\small
\begin{aligned}
\mathbf e_i(0)
=
\cos\theta_i\,\hat{\mathbf x}
+\sin\theta_i\,\hat{\mathbf y},\ 
\mathbf e_i'=
-\frac{
\mathbf r''\!\cdot\!\mathbf e_i
}{
\|\mathbf r'\|^2
}\mathbf r'
+
\frac{\eta'}{\|\mathbf r'\|}
\left(
\mathbf r'\times\mathbf e_i
\right).
\end{aligned}
\label{eq:tendon-direction-transport}
\end{equation}
The first term accounts for the change in centerline direction,
whereas the second represents twist about the tangent. This transport
preserves \(\mathbf e_i\cdot\mathbf r'=0\) and
\(\|\mathbf e_i\|=1\). Hence,
\(\mathbf e_i=\mathbf e_i[\mathbf r,\eta]\) is a derived routing
quantity rather than an additional generalized coordinate. For the straight and untwisted reference configuration,
the current and reference paths of tendon \(i\) are
\begin{equation}
\small
\begin{aligned}
\mathbf p_i(s)=
\mathbf r(s)+\frac{W(s)}{2}\mathbf e_i(s),\ 
\mathbf p_{i0}(s)=
s\hat{\mathbf z}
+\frac{W(s)}{2}\mathbf e_i(0).
\end{aligned}
\label{eq:tendon-paths}
\end{equation}
Differentiating \eqref{eq:tendon-paths} and using
\eqref{eq:tendon-direction-transport} gives the exact current and
reference tendon-path rates
\begin{equation}
\small
\begin{aligned}
\|\mathbf p_i'\|
&=
\Bigg[
\left(
\|\mathbf r'\|
-\frac{W}{2\|\mathbf r'\|}
\mathbf r''\!\cdot\!\mathbf e_i
\right)^2
+\frac{W'^2}{4}
+\frac{W^2}{4}\eta'^2
\Bigg]^{1/2},\\
\|\mathbf p_{i0}'\|
&=
\left(
1+\frac{W'^2}{4}
\right)^{1/2}.
\end{aligned}
\label{eq:tendon-path-rates}
\end{equation}
The effective displacement of tendon \(i\), defined as its current
routed length minus its stress-free routed length, is
\begin{equation}
\small
\begin{aligned}
\Lambda_i[\mathbf r,\eta]
&=
\int_0^L
\chi_i(s)
\left[
\|\mathbf p_i'(s)\|
-\|\mathbf p_{i0}'(s)\|
\right]\mathrm ds .
\end{aligned}
\label{eq:exact-tendon-displacement}
\end{equation}
\subsection{Approximation and Variational Equilibrium}

To isolate the contribution of the routing-diameter gradient, define
\begin{equation}
\small
\begin{aligned}
\varepsilon(s)
=
\frac{W'^2(s)}{4},\ 
\mathcal R_i[\mathbf r,\eta]
=
\Bigg[
\left(
\|\mathbf r'\|
-\frac{W}{2\|\mathbf r'\|}
\mathbf r''\!\cdot\!\mathbf e_i
\right)^2
+\frac{W^2}{4}\eta'^2
\Bigg]^{1/2}.
\end{aligned}
\label{eq:reduced-tendon-rate}
\end{equation}
Here, \(\varepsilon\) is the contribution induced by the
routing-diameter gradient, whereas \(\mathcal R_i\) is the tendon-path
rate excluding this contribution. Eq.
\eqref{eq:tendon-path-rates} becomes
\begin{equation*}
\small
\begin{aligned}
\|\mathbf p_i'\|
&=
\sqrt{\mathcal R_i^2+\varepsilon},
&
\|\mathbf p_{i0}'\|
&=
\sqrt{1+\varepsilon}.
\end{aligned}
\end{equation*}
Parallel tendon routing, widely used in continuum
robots~\cite{childs2021leveraging}, corresponds to \(W'=0\)
in the straight reference configuration considered here.
For longitudinal routing with a small radial slope,
\(\|W'\|_\infty/2\ll1\), \(\varepsilon\) is uniformly small.
We consider regular offset paths satisfying
\begin{equation}
\small
\begin{aligned}
\|\mathbf r'\|
-\frac{W}{2}
\Bigg[
\frac{\mathbf r''\!\cdot\!\mathbf r''}
{\|\mathbf r'\|^2}
-
\frac{
\left(
\mathbf r'\!\cdot\!\mathbf r''
\right)^2
}{
\|\mathbf r'\|^4
}
\Bigg]^{1/2}
&\geq c>0,
\end{aligned}
\label{eq:offset-path-regularity}
\end{equation}
where \(c\) is independent of \(W'\). Since
\begin{equation*}
\small
\begin{aligned}
\left|
\frac{\mathbf r''\!\cdot\!\mathbf e_i}
{\|\mathbf r'\|}
\right|
&\leq
\Bigg[
\frac{\mathbf r''\!\cdot\!\mathbf r''}
{\|\mathbf r'\|^2}
-
\frac{
\left(
\mathbf r'\!\cdot\!\mathbf r''
\right)^2
}{
\|\mathbf r'\|^4
}
\Bigg]^{1/2},
\end{aligned}
\end{equation*}
condition~\eqref{eq:offset-path-regularity} implies
\begin{equation}
\small
\begin{aligned}
\mathcal R_i(s)&\geq c,\\
0\leq
\sup_{s\in[0,L]}
\frac{\varepsilon(s)}{\mathcal R_i^2(s)}
&\leq
\frac{\|W'\|_\infty^2}{4c^2}
\longrightarrow 0
\ 
\text{as }\|W'\|_\infty\longrightarrow0 .
\end{aligned}
\label{eq:uniform-expansion-condition}
\end{equation}
Consequently, uniformly in \(s\),
\begin{equation*}
\small
\begin{aligned}
\|\mathbf p_i'\|-\|\mathbf p_{i0}'\|
&=
\mathcal R_i-1
+\frac{\varepsilon}{2}
\left(
\frac{1}{\mathcal R_i}-1
\right)
+\mathcal O\!\left(\|W'\|_\infty^4\right).
\end{aligned}
\end{equation*}
Substitution into \eqref{eq:exact-tendon-displacement} gives
\begin{equation}
\small
\begin{aligned}
\Lambda_i[\mathbf r,\eta]
&=
\int_0^L
\chi_i(s)
\left[
\mathcal R_i(s)-1
\right]
\,\mathrm ds\\
&\quad+
\frac{1}{2}
\int_0^L
\chi_i(s)\varepsilon(s)
\left[
\frac{1}{\mathcal R_i(s)}-1
\right]
\,\mathrm ds
+\mathcal O\!\left(\|W'\|_\infty^4\right)\\
&=
\int_0^L
\chi_i(s)
\left[
\mathcal R_i(s)-1
\right]
\,\mathrm ds
+\mathcal O\!\left(\|W'\|_\infty^2\right).
\end{aligned}
\label{eq:leading-tendon-displacement}
\end{equation}
The proposed model retains the leading term in
\eqref{eq:leading-tendon-displacement}. Let \(F_i\geq0\) denote the prescribed tension of tendon \(i\).
Neglecting tendon friction makes \(F_i\) constant along its routed
path. Using \eqref{eq:leading-tendon-displacement}, the leading-order total potential energy incorporates tendon work in actuation space \cite{wu2026lightweight,yang2026lightweightC}:
\begin{equation}
\small
\begin{aligned}
\Pi[\mathbf r,\eta]
&=
\frac{1}{2}
\int_0^L
\Bigg[
EA(s)
\left(
\|\mathbf r'\|-1
\right)^2+EI(s)
\left\|
\frac{\mathrm d}{\mathrm ds}
\left(
\frac{\mathbf r'}{\|\mathbf r'\|}
\right)
\right\|^2\\
&\quad+GJ(s)\eta'^2
\Bigg]\mathrm ds+\sum_{i=1}^{n}F_i
\int_0^L
\chi_i(s)
\left[
\mathcal R_i[\mathbf r,\eta]-1
\right]\mathrm ds .
\end{aligned}
\label{eq:leading-total-potential}
\end{equation}
Let \(\delta\boldsymbol{\phi}\) denote the infinitesimal rotation
induced by variations of \(\mathbf r\) and \(\eta\). The
required identities are
\begin{equation}
\small
\centering
\begin{aligned}
&\delta\mathbf r'
=
\frac{\mathbf r'}{\|\mathbf r'\|}
\delta\|\mathbf r'\|
+
\delta\boldsymbol{\phi}\times\mathbf r',\ 
\delta\eta'
=
\frac{\mathbf r'}{\|\mathbf r'\|}
\cdot\delta\boldsymbol{\phi}',\\
&\delta
\left\|
\frac{\mathrm d}{\mathrm ds}
\left(
\frac{\mathbf r'}{\|\mathbf r'\|}
\right)
\right\|^2
=
2
\frac{
\mathbf r'\times\mathbf r''
}{
\|\mathbf r'\|^2
}
\cdot\delta\boldsymbol{\phi}',\ 
\delta
\left(
\frac{
\mathbf r''\!\cdot\!\mathbf e_i
}{
\|\mathbf r'\|
}
\right)
=
\frac{
\mathbf r'\times\mathbf e_i
}{
\|\mathbf r'\|}
\cdot\delta\boldsymbol{\phi}'.
\end{aligned}
\label{eq:variational-identities}
\end{equation}
Since \(W(s)\) is prescribed, \(\delta W=0\). Applying
\eqref{eq:variational-identities} to
\eqref{eq:reduced-tendon-rate} gives
\begin{equation}
\small
\begin{aligned}
\delta\mathcal R_i
&=
\frac{1}{\mathcal R_i}
\left(
\|\mathbf r'\|
-\frac{W}{2\|\mathbf r'\|}
\mathbf r''\!\cdot\!\mathbf e_i
\right)
\delta\|\mathbf r'\|
+\frac{1}{\mathcal R_i}
\Bigg[
-\frac{W}{2\|\mathbf r'\|}
\\
&\times \left(
\|\mathbf r'\|
-\frac{W}{2\|\mathbf r'\|}
\mathbf r''\!\cdot\!\mathbf e_i
\right)
\left(
\mathbf r'\times\mathbf e_i
\right)+\frac{W^2\eta'}{4\|\mathbf r'\|}
\mathbf r'
\Bigg]
\cdot\delta\boldsymbol{\phi}'.
\end{aligned}
\label{eq:reduced-tendon-rate-variation}
\end{equation}
Substituting \eqref{eq:variational-identities} and
\eqref{eq:reduced-tendon-rate-variation} into
\(\delta\Pi=0\) gives
\begin{equation}
\small
\begin{aligned}
\delta\Pi
&=
\int_0^L
\Bigg[
EA(s)
\left(
\|\mathbf r'\|-1
\right)\\
&\quad
+\sum_{i=1}^{n}
\frac{\chi_i(s)F_i}{\mathcal R_i}
\left(
\|\mathbf r'\|
-\frac{W}{2\|\mathbf r'\|}
\mathbf r''\!\cdot\!\mathbf e_i
\right)
\Bigg]
\delta\|\mathbf r'\|\,\mathrm ds\\
&\quad
+\sum_{j=1}^{N_s}
\int_{\mathcal I_j}
\mathbf M(s)\cdot
\delta\boldsymbol{\phi}'\,\mathrm ds ,
\end{aligned}
\label{eq:first-variation}
\end{equation}
where the total bending--torsional moment is
\begin{equation}
\small
\begin{aligned}
\mathbf M
&=
\frac{EI(s)}{\|\mathbf r'\|^2}
\left(
\mathbf r'\times\mathbf r''
\right)
+\frac{GJ(s)\eta'}{\|\mathbf r'\|}
\mathbf r'\\
&\quad
+\sum_{i=1}^{n}
\frac{\chi_i(s)F_i}{\mathcal R_i}
\Bigg[
-\frac{W}{2\|\mathbf r'\|}
\left(
\|\mathbf r'\|
-\frac{W}{2\|\mathbf r'\|}
\mathbf r''\!\cdot\!\mathbf e_i
\right)\\
&\quad
\times\left(
\mathbf r'\times\mathbf e_i
\right)+\frac{W^2\eta'}{4\|\mathbf r'\|}
\mathbf r'
\Bigg].
\end{aligned}
\label{eq:total-moment}
\end{equation}
Integrating the rotational term in \eqref{eq:first-variation} by parts
over each segment gives
\begin{equation*}
\small
\begin{aligned}
\sum_{j=1}^{N_s}
\int_{\mathcal I_j}
\mathbf M\cdot\delta\boldsymbol{\phi}'\,\mathrm ds
=
\sum_{j=1}^{N_s}
\left[
\mathbf M\cdot\delta\boldsymbol{\phi}
\right]_{s_{j-1}}^{s_j}
-\sum_{j=1}^{N_s}
\int_{\mathcal I_j}
\mathbf M'\cdot\delta\boldsymbol{\phi}\,\mathrm ds .
\end{aligned}
\end{equation*}
Since the base is clamped,
\(\delta\boldsymbol{\phi}(0)=\mathbf0\). Stationarity for arbitrary
segment-wise, interface, and free-tip variations requires
\(\mathbf M'=\mathbf0\) in each segment, continuity of \(\mathbf M\)
across every interface, and \(\mathbf M(L)=\mathbf0\). Consequently,
\(\mathbf M=\mathbf0\) throughout the backbone. Independently, the
coefficient of the arbitrary axial variation
\(\delta\|\mathbf r'\|\) must vanish. The equilibrium conditions are
therefore
\begin{equation}
\small
\begin{aligned}
&\mathbf M(s)
=\mathbf0,\\
&EA(s)
\left(
\|\mathbf r'\|-1
\right)+\sum_{i=1}^{n}
\frac{\chi_i(s)F_i}{\mathcal R_i}
\left(
\|\mathbf r'\|
-\frac{W}{2\|\mathbf r'\|}
\mathbf r''\!\cdot\!\mathbf e_i
\right)
=0.
\end{aligned}
\label{eq:variational-equilibrium}
\end{equation}
Projecting \(\mathbf M (s)=\mathbf0\) onto
\(\mathbf r'/\|\mathbf r'\|\) eliminates all bending contributions and
gives
\begin{equation}
\small
\begin{aligned}
0
&=
\frac{\mathbf r'}{\|\mathbf r'\|}
\cdot\mathbf M=
\eta'
\Bigg[
GJ(s)
+
\sum_{i=1}^{n}
\frac{
\chi_i(s)F_iW^2(s)
}{
4\mathcal R_i
}
\Bigg].
\end{aligned}
\label{eq:twist-equilibrium}
\end{equation}
For \(GJ(s)>0\), \(F_i\geq0\), and
\(\mathcal R_i\geq c>0\), the bracket is strictly positive. Hence,
\(\eta'(s)=0\); together with \(\eta(0)=0\),
\begin{equation*}
\small
\begin{aligned}
\eta(s)=0,
\ s\in[0,L].
\end{aligned}
\end{equation*}
With \(\eta'=0\), the offset-path regularity condition makes
\[
\mathcal R_i
=
\|\mathbf r'\|
-\frac{W}{2\|\mathbf r'\|}
\mathbf r''\!\cdot\!\mathbf e_i
>0.
\]
Substitution into \eqref{eq:variational-equilibrium} and
\(\mathbf M=\mathbf0\) yields the direct Cartesian
force-to-configuration relations
\begin{equation}
\small
{
\begin{aligned}
&\eta(s)
=0,\ \|\mathbf r'(s)\|
=
1-
\frac{
\displaystyle\sum_{i=1}^{n}\chi_i(s)F_i
}{
EA(s)
},\\
&EI(s)
\frac{\mathrm d}{\mathrm ds}
\left(
\frac{\mathbf r'}{\|\mathbf r'\|}
\right)
=
\frac{W(s)}{2}
\sum_{i=1}^{n}
\chi_i(s)F_i\mathbf e_i(s).
\end{aligned}}
\label{eq:direct-force-to-configuration}
\end{equation}
Eq.~\eqref{eq:direct-force-to-configuration} is valid while the
offset tendon paths remain regular and
\(\sum_i\chi_i(s)F_i<EA(s)\), which guarantees
\(\|\mathbf r'(s)\|>0\).

\subsection{Closed-Form Cartesian Configuration}

Within segment \(\mathcal I_j\), the active tendon set
\(\Gamma_j\) is fixed. Define the transverse tendon-force resultant
\begin{equation}
\small
\begin{aligned}
\mathbf f_j(s)
&=
\sum_{i\in\Gamma_j}
F_i\mathbf e_i(s),\\
\|\mathbf f_j\|
&=
\Bigg[
\left(
\sum_{i\in\Gamma_j}F_i\cos\theta_i
\right)^2
+
\left(
\sum_{i\in\Gamma_j}F_i\sin\theta_i
\right)^2
\Bigg]^{1/2}.
\end{aligned}
\label{eq:segment-tendon-resultant}
\end{equation}
Under the zero-twist transport,
\(\mathrm d(\mathbf e_i\cdot\mathbf e_k)/\mathrm ds=0\);
hence, the relative tendon angles and
\(\|\mathbf f_j\|\) remain constant within the segment. For \(s\in\mathcal I_j\),
\eqref{eq:direct-force-to-configuration} reduces to
\begin{equation}
\small
\begin{aligned}
\|\mathbf r'(s)\|
=
1-
\frac{
\displaystyle\sum_{i\in\Gamma_j}F_i
}{
EA(s)
},\ 
\frac{\mathrm d}{\mathrm ds}
\left(
\frac{\mathbf r'}{\|\mathbf r'\|}
\right)
=
\frac{W(s)}
{2EI(s)}
\mathbf f_j(s).
\end{aligned}
\label{eq:segment-force-to-configuration}
\end{equation}
Summing the zero-twist transport equation over the active tendons and
using \eqref{eq:segment-force-to-configuration} gives
\begin{equation*}
\small
\begin{aligned}
\mathbf f_j'
&=
-\frac{W(s)}
{2EI(s)}
\|\mathbf f_j\|^2
\frac{\mathbf r'}{\|\mathbf r'\|}.
\end{aligned}
\end{equation*}
For \(\|\mathbf f_j\|>0\), define the unit normal to the segment
bending plane as
\begin{equation}
\small
\begin{aligned}
\mathbf b_j
=
\frac{
\mathbf r'\times\mathbf f_j
}{
\|\mathbf r'\|\,\|\mathbf f_j\|
},\ 
\mathbf b_j'=
\frac{1}{\|\mathbf f_j\|}
\Bigg[
\frac{\mathrm d}{\mathrm ds}
\left(
\frac{\mathbf r'}{\|\mathbf r'\|}
\right)
\times\mathbf f_j
+
\frac{\mathbf r'}{\|\mathbf r'\|}
\times\mathbf f_j'
\Bigg]
=\mathbf0.
\end{aligned}
\label{eq:fixed-segment-bending-plane}
\end{equation}
If \(\|\mathbf f_j\|=0\), the segment remains straight but may
undergo axial deformation. In this case,
\begin{equation}
\small
\begin{aligned}
\mathbf r(s)
&=
\mathbf r(s_{j-1})
+
\frac{\mathbf r'(s_{j-1})}
{\|\mathbf r'(s_{j-1})\|}
\int_{s_{j-1}}^{s}
\left[
1-
\frac{\sum_{i\in\Gamma_j}F_i}
{EA(\xi)}
\right]\mathrm d\xi,\\
\mathbf e_i(s)
&=\mathbf e_i(s_{j-1}),
\qquad s\in\mathcal I_j .
\end{aligned}
\label{eq:zero-resultant-segment}
\end{equation}
For \(\|\mathbf f_j\|>0\), define the accumulated bending angle within
segment
\(\mathcal I_j\) as
\begin{equation}
\small
\begin{aligned}
\Phi_j(s)
&=
\|\mathbf f_j\|
\int_{s_{j-1}}^{s}
\frac{W(\xi)}
{2EI(\xi)}
\,\mathrm d\xi .
\end{aligned}
\label{eq:segment-bending-angle}
\end{equation}
Since \(\mathbf b_j\) is constant,
\eqref{eq:segment-force-to-configuration} integrates to
\begin{equation}
\small
\begin{aligned}
\frac{\mathbf r'(s)}
{\|\mathbf r'(s)\|}
=
\cos\Phi_j(s)
\frac{\mathbf r'(s_{j-1})}
{\|\mathbf r'(s_{j-1})\|}+\sin\Phi_j(s)
\left[
\mathbf b_j\times
\frac{\mathbf r'(s_{j-1})}
{\|\mathbf r'(s_{j-1})\|}
\right],
\\ s\in\mathcal I_j.
\end{aligned}
\label{eq:closed-form-tangent}
\end{equation}
Combining \eqref{eq:segment-force-to-configuration} and
\eqref{eq:closed-form-tangent} yields the segment centerline
\begin{equation}
\small
\begin{aligned}
\mathbf r(s)
&=
\mathbf r(s_{j-1})+\frac{\mathbf r'(s_{j-1})}
{\|\mathbf r'(s_{j-1})\|}
\int_{s_{j-1}}^{s}
\left[
1-
\frac{
\displaystyle\sum_{i\in\Gamma_j}F_i
}{
EA(\xi)
}
\right]
\cos\Phi_j(\xi)\,\mathrm d\xi\\
&\ 
+\left[
\mathbf b_j\times
\frac{\mathbf r'(s_{j-1})}
{\|\mathbf r'(s_{j-1})\|}
\right]
\int_{s_{j-1}}^{s}\!\!
\left[
1-
\frac{
\displaystyle\sum_{i\in\Gamma_j}F_i
}{
EA(\xi)
}
\right]
\sin\Phi_j(\xi)\,\mathrm d\xi,
\end{aligned}
\label{eq:closed-form-centerline}
\end{equation}
for \(s\in\mathcal I_j\). For \(\|\mathbf f_j\|>0\), the tendon
directions undergo the same segment rotation:
\begin{equation}
\small
\begin{aligned}
\mathbf e_i(s)
&=
\cos\Phi_j(s)\,
\mathbf e_i(s_{j-1})+\sin\Phi_j(s)
\left[
\mathbf b_j\times\mathbf e_i(s_{j-1})
\right]\\
&\quad
+\left[
1-\cos\Phi_j(s)
\right]
\left[
\mathbf b_j\cdot\mathbf e_i(s_{j-1})
\right]\mathbf b_j,
\  s\in\mathcal I_j.
\end{aligned}
\label{eq:closed-form-tendon-direction}
\end{equation}
Starting from the base conditions in
\eqref{eq:cartesian-generalized-coordinate} and
\eqref{eq:tendon-direction-transport}, the solution is propagated
segment by segment. For each \(j=1,\ldots,N_s\), the tendon-force
resultant \(\mathbf f_j(s_{j-1})\) is first evaluated from
\eqref{eq:segment-tendon-resultant}. If
\(\|\mathbf f_j(s_{j-1})\|>0\), the bending-plane normal
\(\mathbf b_j\) is obtained from
\eqref{eq:fixed-segment-bending-plane}, and the segment is propagated
using \eqref{eq:closed-form-tangent},
\eqref{eq:closed-form-centerline}, and
\eqref{eq:closed-form-tendon-direction}. If
\(\|\mathbf f_j(s_{j-1})\|=0\), the straight-segment solution
\eqref{eq:zero-resultant-segment} is used instead. In either case, the
values at \(s=s_j\) provide the proximal data for segment \(j+1\). Consequently, the applied tendon forces determine the Cartesian
generalized-coordinate field directly as
\begin{equation*}
\small
\begin{aligned}
\{F_i\}_{i=1}^{n}
\quad\longmapsto\quad
\mathbf q(s)
&=
\begin{bmatrix}
\mathbf r^{\mathbf T}(s)&\eta(s)
\end{bmatrix}^{\mathbf T}
=
\begin{bmatrix}
\mathbf r^{\mathbf T}(s)&0
\end{bmatrix}^{\mathbf T}.
\end{aligned}
\label{eq:closed-form-force-to-configuration}
\end{equation*}
The resulting leading-order solution is available in quadratures for
spatially varying stiffness and routing profiles \(EA(s)\), \(EI(s)\),
\(GJ(s)\), and \(W(s)\), provided that the tendon-path regularity and
axial-deformation conditions are satisfied. The underlying tendon-displacement approximation neglects terms of order \(\mathcal O(\|W'\|_\infty^2)\).

\section{Numerical Validation}
\label{sec:numerical-validation}

The numerical study evaluates two aspects of the proposed model: its prediction of Cartesian configurations and distributed strains, and the computational cost of repeated force-to-configuration evaluations. Both single- and three-segment robots are compared with a full-strain numerical reference.

\subsection{Simulation Setup}
\label{subsec:simulation-setup}

The proposed formulation is evaluated against numerical reference solutions obtained using GVS \cite{renda2020geometric}. The common physical and numerical parameters and the two robot
configurations are summarized in
Table~\ref{tab:simulation-parameters}. In particular, the tabulated
values of \(N_s\), \(\{s_j\}_{j=0}^{N_s}\), and
\(\{\Gamma_j\}_{j=1}^{N_s}\) instantiate the segment construction in
Sec.~\ref{subsec:generalized-coordinates} and uniquely determine the
intervals \(\mathcal I_j\) and routing indicators \(\chi_i(s)\). The two robots share the same material properties, cross-sectional
and tendon-routing profiles, and numerical settings.  For the GVS reference model, the strain field is represented using
fifth-order polynomial basis functions. Static equilibria are computed with
a residual acceptance threshold of \(10^{-7}\), and each load step is
warm-started from the converged solution of the preceding step to improve
numerical robustness.

\begin{table}[!t]
\centering
\caption{Parameters, numerical settings, and robot
configurations.}
\label{tab:simulation-parameters}
\footnotesize
\setlength{\tabcolsep}{2pt}
\renewcommand{\arraystretch}{1.05}

\begin{tabular}{
@{}
p{0.36\columnwidth}
p{0.12\columnwidth}
p{0.40\columnwidth}
p{0.06\columnwidth}
@{}}
\toprule
\multicolumn{4}{c}{\textbf{Physical and geometric parameters}} \\
\midrule
\textbf{Parameter}
& \textbf{Symbol}
& \textbf{Value or setting}
& \textbf{Unit} \\
\midrule

Backbone length
& \(L\)
& 0.4
& m \\

Young's modulus
& \(E\)
& $1\times10^7$
& Pa \\

Poisson's ratio
& \(\nu_{\mathrm p}\)
& 0.49
& -- \\

Shear modulus
& \(G\)
& \(\displaystyle\frac{E}{2(1+\nu_{\mathrm p})}\)
& Pa \\

Backbone diameter
& \(D(s)\)
& $4.0\times10^{-3}(1-0.15s/L)$
& m \\

Cross-sectional area
& \(A(s)\)
& \(\pi D^2(s)/4\)
& \(\mathrm{m^2}\) \\

Second moment of area
& \(I(s)\)
& \(\pi D^4(s)/64\)
& \(\mathrm{m^4}\) \\

Torsional constant
& \(J(s)\)
& \(\pi D^4(s)/32\)
& \(\mathrm{m^4}\) \\

Tendon-routing diameter
& \(W(s)\)
& $5\times10^{-3}[1-0.3(s/L)^3]$
& m \\

Number of tendons
& \(n\)
& 3 (single); 9 (three-segment)
& -- \\

Tendon angular positions
& \(\{\theta_i\}_{i=1}^{n}\)
& \(\{0, 2\pi/3,4\pi/3,0,2\pi/3...\ \}\)
& rad \\

Calibrated axial stiffness
& $K_A(s)$
& \( 0.05EA(s)\)
& \(\mathrm{N}\) \\

\midrule
\multicolumn{4}{c}{\textbf{Numerical implementation}} \\
\midrule

Spatial integration nodes
& \(N_d\)
& 181
& -- \\

Computing platform
& --
& Intel Core i9-14900HX
& -- \\

\bottomrule
\end{tabular}

\par\vspace{1.5mm}

\begin{tabular}{
@{}
p{0.40\columnwidth}
p{0.16\columnwidth}
p{0.30\columnwidth}
@{}}
\multicolumn{3}{c}{
\textbf{Single-Segment Robot Configuration}} \\
\midrule
\textbf{Definition}
& \textbf{Symbol}
& \textbf{Setting} \\
\midrule

Number of segments
& \(N_s\)
& \(1\) \\

Segment boundaries
& \(\{s_j\}_{j=0}^{1}\)
& \(\{0,L\}\) \\

Active tendon set
& \(\Gamma_1\)
& \(\{1,2,3\}\) \\

\bottomrule
\end{tabular}

\par\vspace{1.5mm}

\begin{tabular}{
@{}
p{0.40\columnwidth}
p{0.16\columnwidth}
p{0.30\columnwidth}
@{}}
\multicolumn{3}{c}{
\textbf{Three-Segment Robot Configuration}} \\
\midrule
\textbf{Definition}
& \textbf{Symbol}
& \textbf{Setting} \\
\midrule

Number of segments
& \(N_s\)
& \(3\) \\

Segment boundaries
& \(\{s_j\}_{j=0}^{3}\)
& \(\{0,L/3,2L/3,L\}\)\\

Segment-1 tendon set
& \(\Gamma_1\)
& \{1,2,3,4,5,6,7,8,9\}\\

Segment-2 tendon set
& \(\Gamma_2\)
& \{4,5,6,7,8,9\} \\

Segment-3 tendon set
& \(\Gamma_3\)
& \{7,8,9\}\\

\bottomrule
\end{tabular}
\end{table}

\subsection{Cartesian Configuration and Distributed-Strain Accuracy}
\label{subsec:configuration-strain-accuracy}

The full-mode GVS model is used as the numerical reference for the
configuration and strain comparisons. It retains all six Cosserat
strain components, thereby admitting axial extension, transverse
shear, bending, and material twist. Its tendon paths are reconstructed
directly from the cross-sectional poses without applying the
leading-order reduction in
\eqref{eq:leading-tendon-displacement}. The resulting comparisons
therefore evaluate the overall modeling accuracy. The distributed-strain quantities used in the comparisons are
summarized in Table~\ref{tab:strain-comparison}. GVS reports the
strain components in its material frame, whereas the corresponding
quantities of the proposed model are derived directly from
\(\mathbf r\) and \(\eta\). For each robot, three prescribed nonnegative tendon-force vectors are
considered. The single- and three-segment inputs,
\(\mathbf F_{\mathrm S}^{(\ell)}\) and
\(\mathbf F_{\mathrm M}^{(\ell)}\), respectively, are listed together
in Table~\ref{tab:tendon-force-cases}.

\begin{table}[!t]
\centering
\caption{Distributed-strain quantities used for numerical validation.}
\label{tab:strain-comparison}
\scriptsize
\setlength{\tabcolsep}{1.3pt}
\renewcommand{\arraystretch}{1.10}
\begin{tabular}{
@{}
p{0.34\columnwidth}
p{0.22\columnwidth}
p{0.22\columnwidth}
p{0.18\columnwidth}
@{}}
\toprule
\textbf{Strain measure}
& \textbf{GVS}
& \textbf{Proposed}
& \textbf{Unit} \\
\midrule

Transverse shear
& \((\nu_1,\nu_2)\)
& \((0,0)\)
& -- \\

Axial strain
& \(\nu_3-1\)
& \(\|\mathbf r'\|-1\)
& -- \\

Bending magnitude ($\kappa_b$)
& \(\sqrt{\kappa_1^2+\kappa_2^2}\)
& \(\displaystyle
\left\|
\left(
\frac{\mathbf r'}{\|\mathbf r'\|}
\right)'
\right\|\)
& \(\mathrm{m^{-1}}\) \\

Torsional strain
& \(\kappa_3\)
& \(\eta'=0\)
& \(\mathrm{m^{-1}}\) \\

\bottomrule
\end{tabular}
\end{table}


\begin{table}[!t]
\centering
\caption{\scriptsize Tendon-force cases for the single- and three-segment robots.}
\label{tab:tendon-force-cases}

\footnotesize
\setlength{\tabcolsep}{2pt}
\setlength{\arraycolsep}{2pt}
\renewcommand{\arraystretch}{1.08}

\begin{tabularx}{\columnwidth}{
@{}
c
>{\centering\arraybackslash}p{0.28\columnwidth}
>{\centering\arraybackslash}X
c
@{}
}
\toprule
\textbf{Case}
& \textbf{Single-segment robot}
& \textbf{Three-segment robot}
& \textbf{Unit} \\
\midrule

1
&
\(
\mathbf{F}_{\mathrm{S}}^{(1)}=
\begin{bmatrix}
0.4288\\
0.2800\\
0.2464
\end{bmatrix}
\)
&
\(
\mathbf{F}_{\mathrm{M}}^{(1)}=
\begin{bmatrix}
0.16288 & 0.12800 & 0.12864\\
0.16032 & 0.19440 & 0.04032\\
0.18432 & 0.02432 & 0.18608
\end{bmatrix}
\)
& N\\
\addlinespace[6pt]

2
&
\(
\mathbf{F}_{\mathrm{S}}^{(2)}=
\begin{bmatrix}
0.0864\\
0.3728\\
0.1600
\end{bmatrix}
\)
&
\(
\mathbf{F}_{\mathrm{M}}^{(2)}=
\begin{bmatrix}
0.02080 & 0.10360 & 0.02000\\
0.32040 & 0.10040 & 0.24300\\
0.02040 & 0.19040 & 0.27260
\end{bmatrix}
\)
& N\\
\addlinespace[6pt]

3
&
\(
\mathbf{F}_{\mathrm{S}}^{(3)}=
\begin{bmatrix}
0.2100\\
0.2156\\
0.3752
\end{bmatrix}
\)
&
\(
\mathbf{F}_{\mathrm{M}}^{(3)}=
\begin{bmatrix}
0.17500 & 0.10556 & 0.00252\\
0.09828 & 0.24010 & 0.02828\\
0.02828 & 0.20482 & 0.20328
\end{bmatrix}
\)
& N\\[3pt]

\bottomrule
\end{tabularx}
\end{table}

\subsubsection{Single-Segment Case}
\label{subsubsec:single-segment-case}
\begin{figure*}[!t]
\centering
\captionsetup[subfloat]{font=scriptsize}

\subfloat[\(\mathbf F_{\mathrm S}^{(1)}\): backbone
\label{fig:single-case1-backbone}]
{\includegraphics[width=0.18\textwidth]
{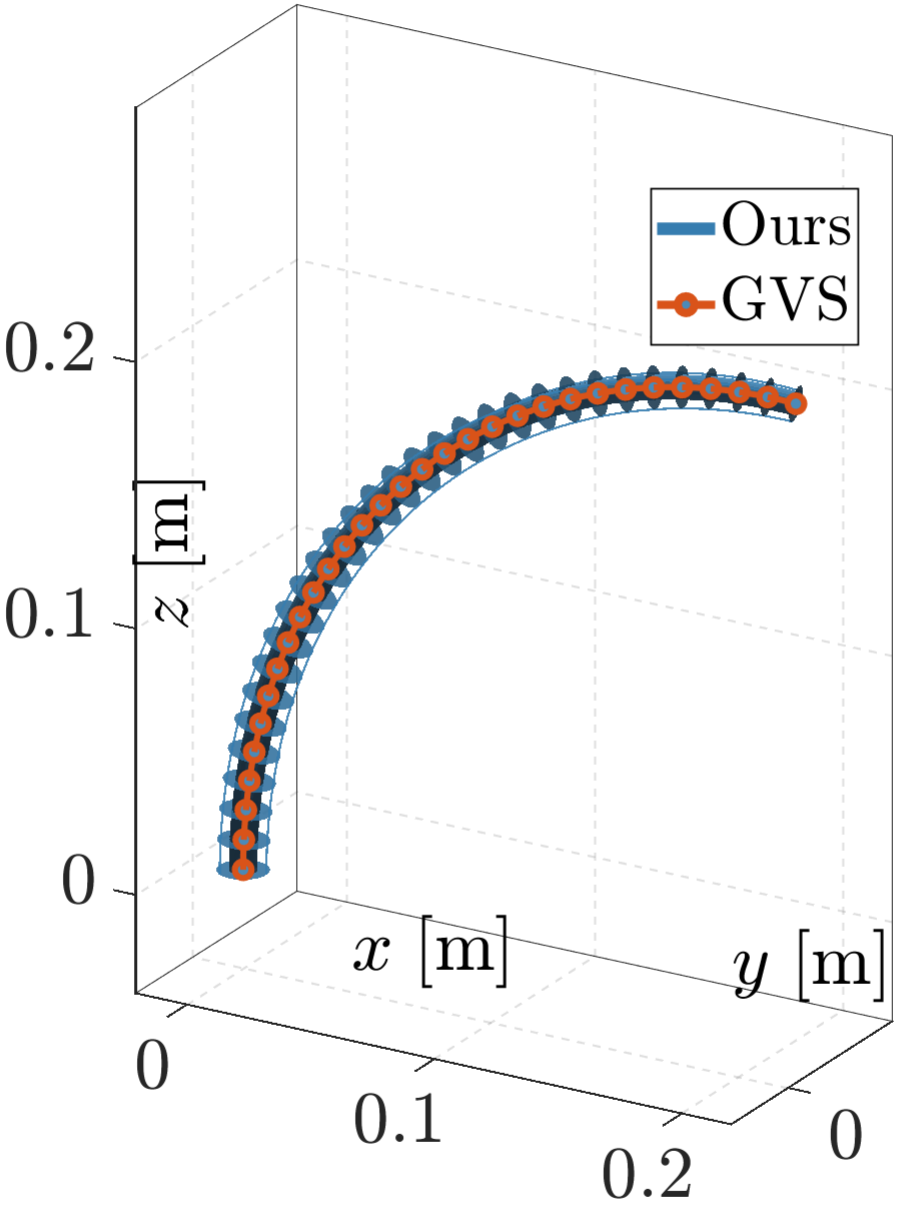}}
\hfill
\subfloat[\(\mathbf F_{\mathrm S}^{(1)}\): \(x,y,z\)
\label{fig:single-case1-cartesian}]
{\includegraphics[width=0.246\textwidth]
{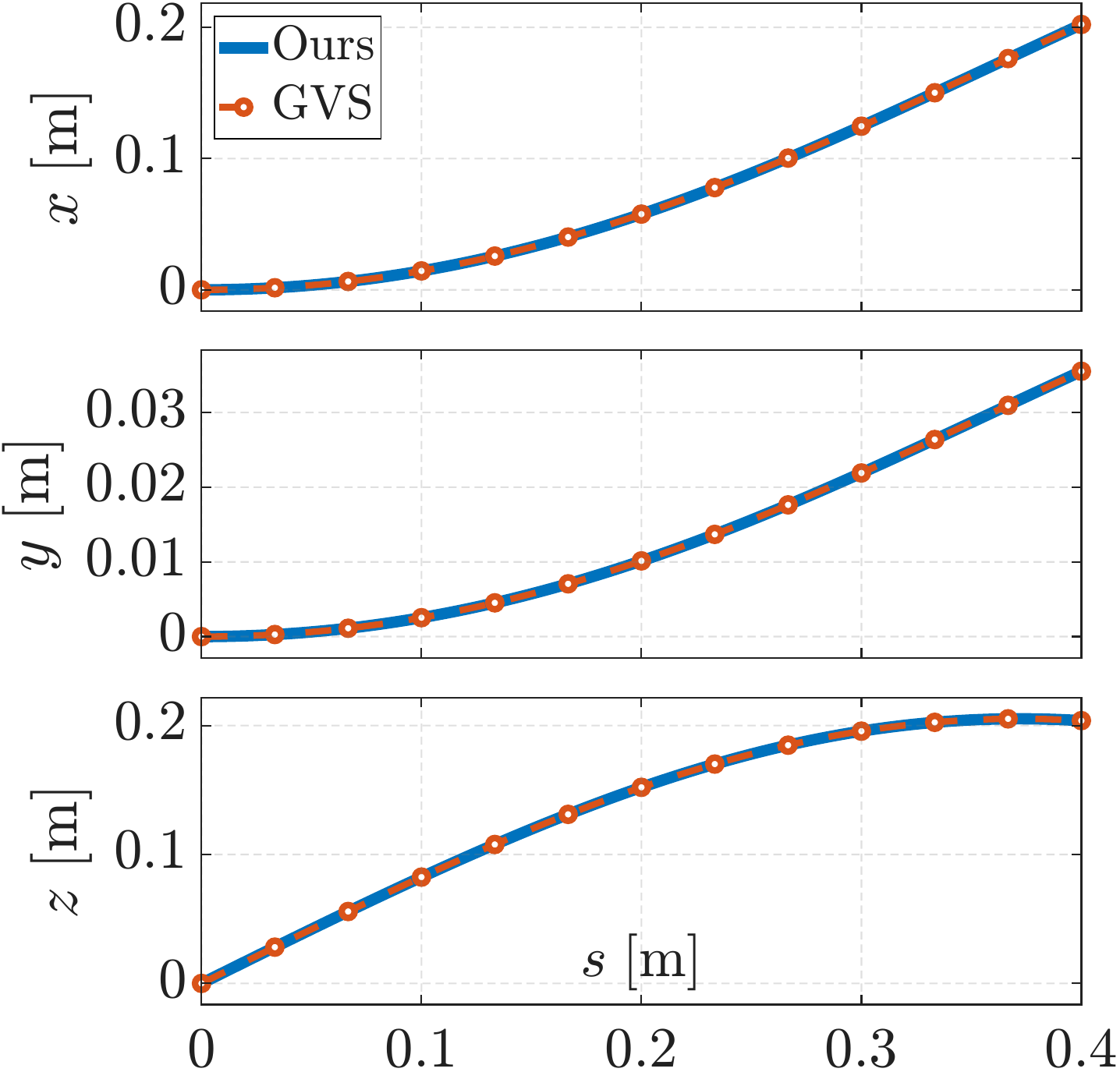}}
\hfill
\subfloat[\(\mathbf F_{\mathrm S}^{(1)}\): \(\boldsymbol{\nu}\)
\label{fig:single-case1-nu}]
{\includegraphics[width=0.242\textwidth]
{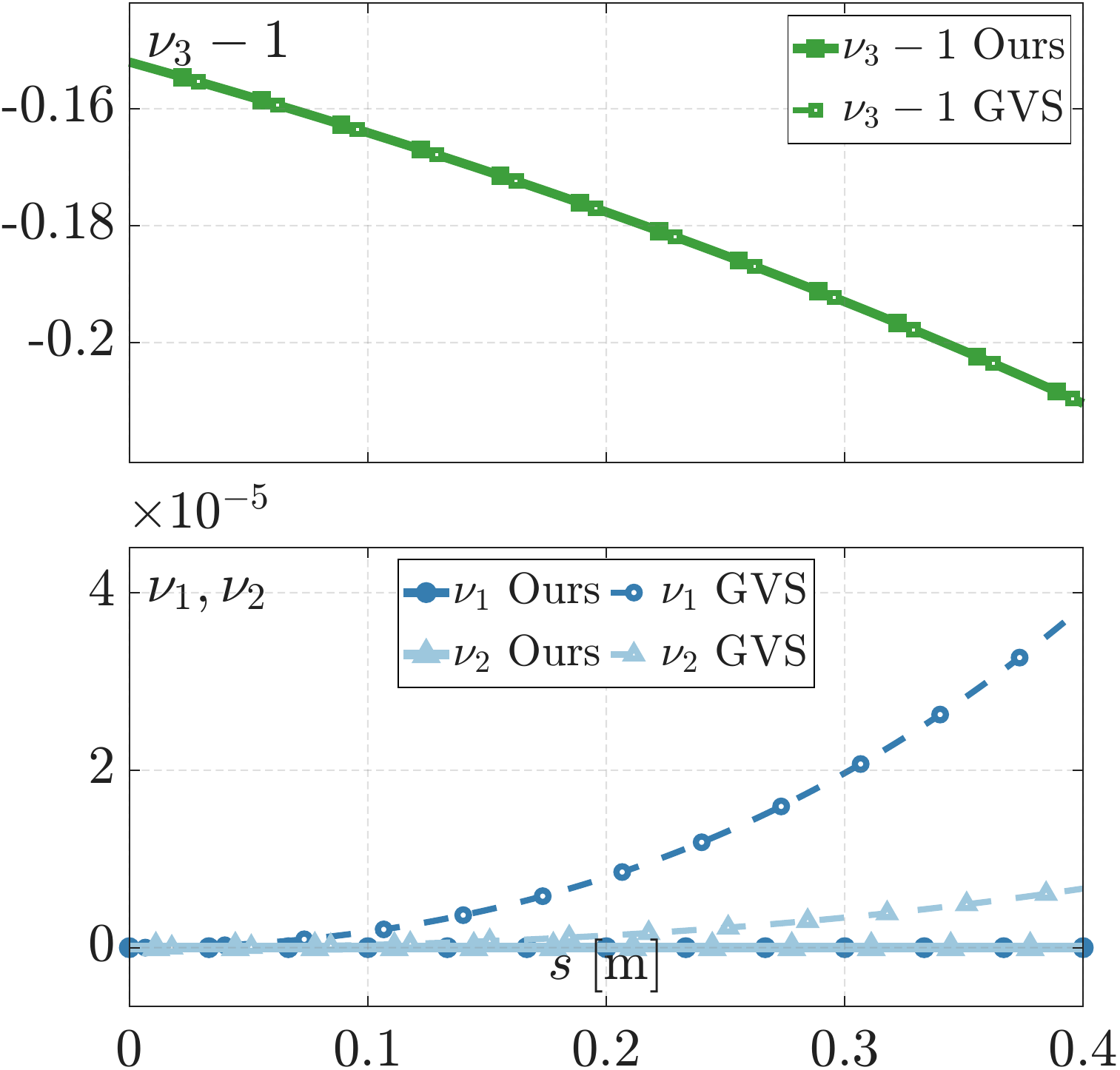}}
\hfill
\subfloat[\(\mathbf F_{\mathrm S}^{(1)}\): \(\kappa_b,\kappa_3\)
\label{fig:single-case1-kappa}]
{\includegraphics[width=0.235\textwidth]
{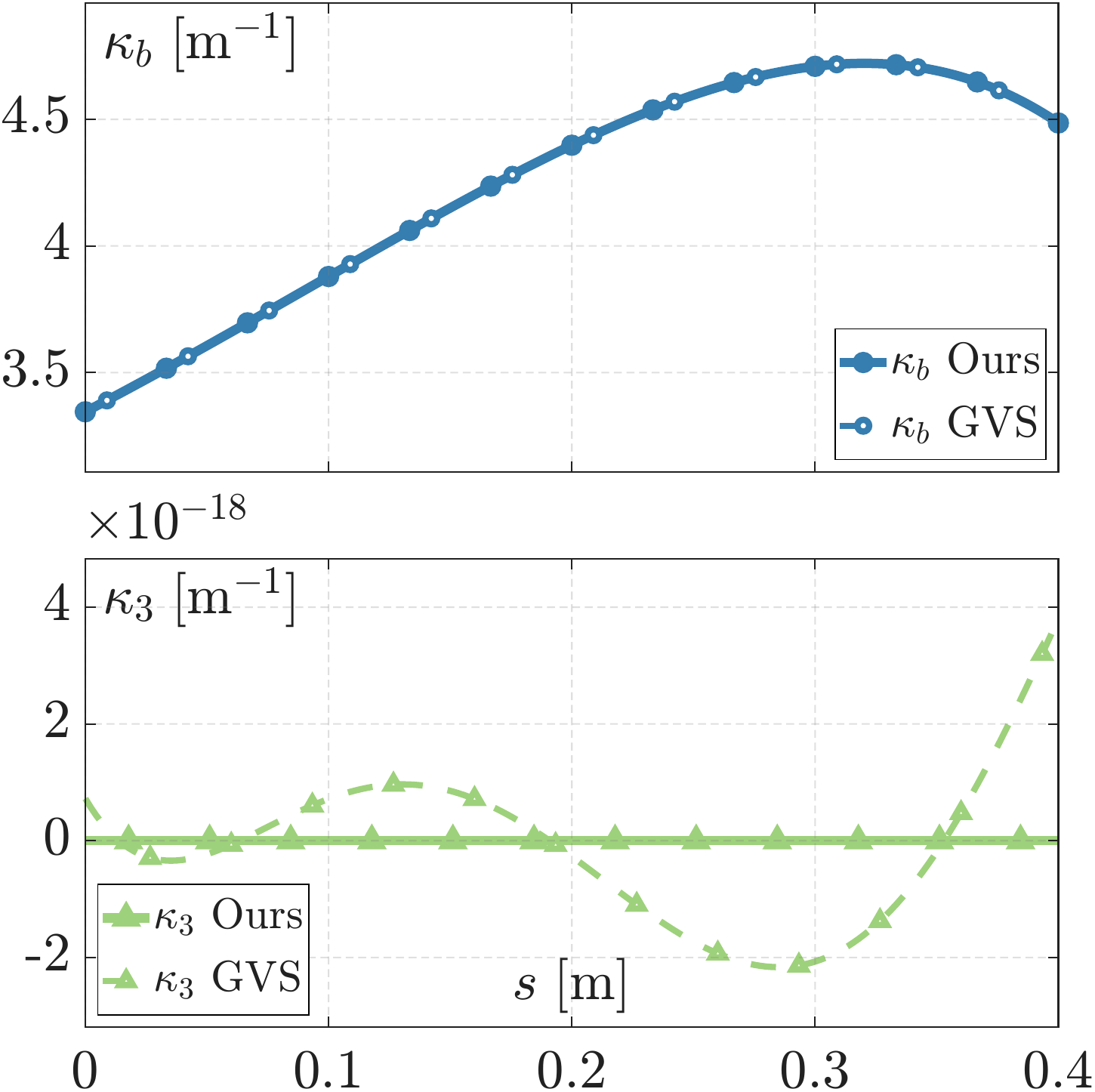}}

\par\vspace{-1.8mm}

\subfloat[\(\mathbf F_{\mathrm S}^{(2)}\): backbone
\label{fig:single-case2-backbone}]
{\includegraphics[width=0.22\textwidth]
{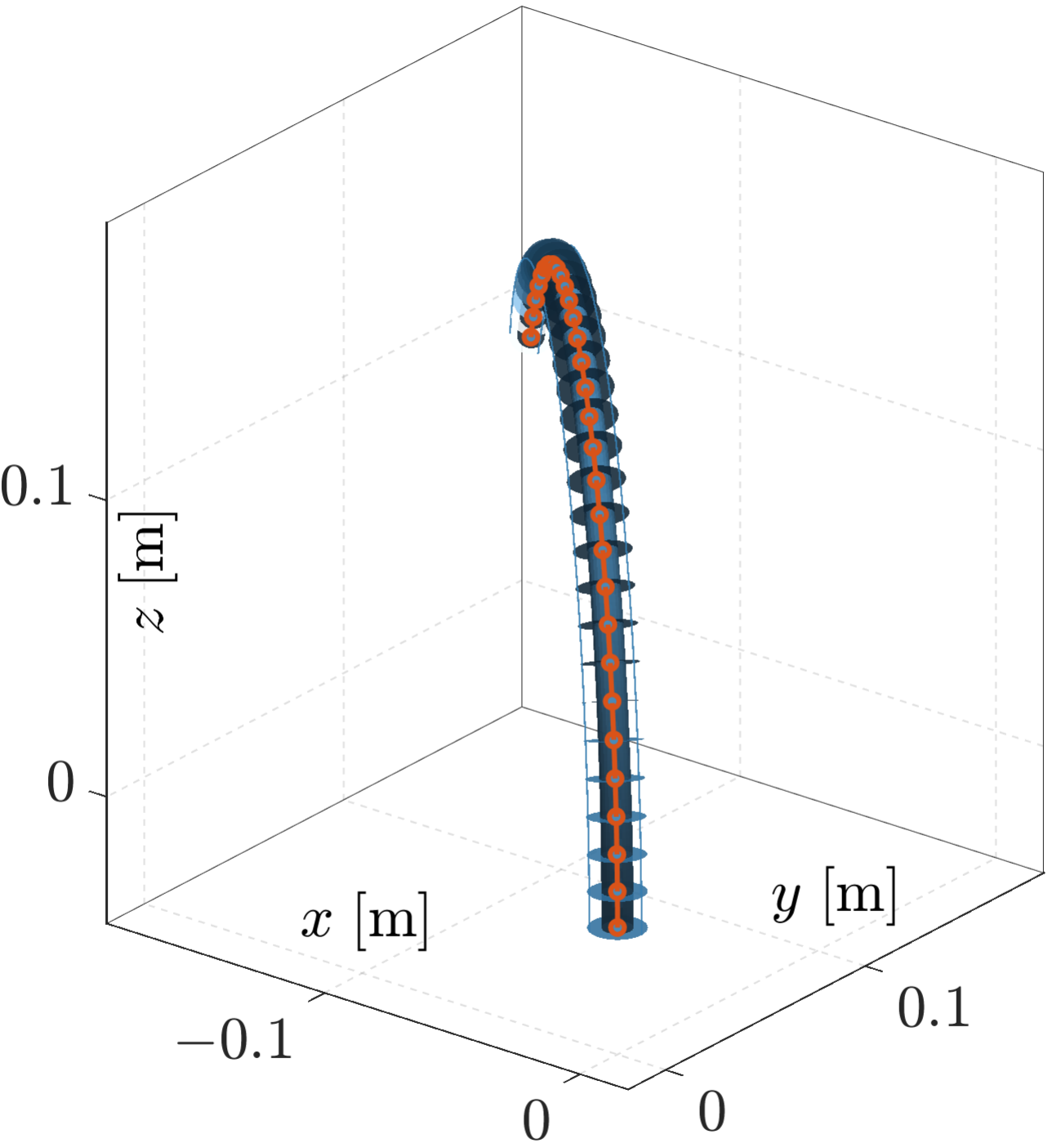}}
\hfill
\subfloat[\(\mathbf F_{\mathrm S}^{(2)}\): \(x,y,z\)
\label{fig:single-case2-cartesian}]
{\includegraphics[width=0.245\textwidth]
{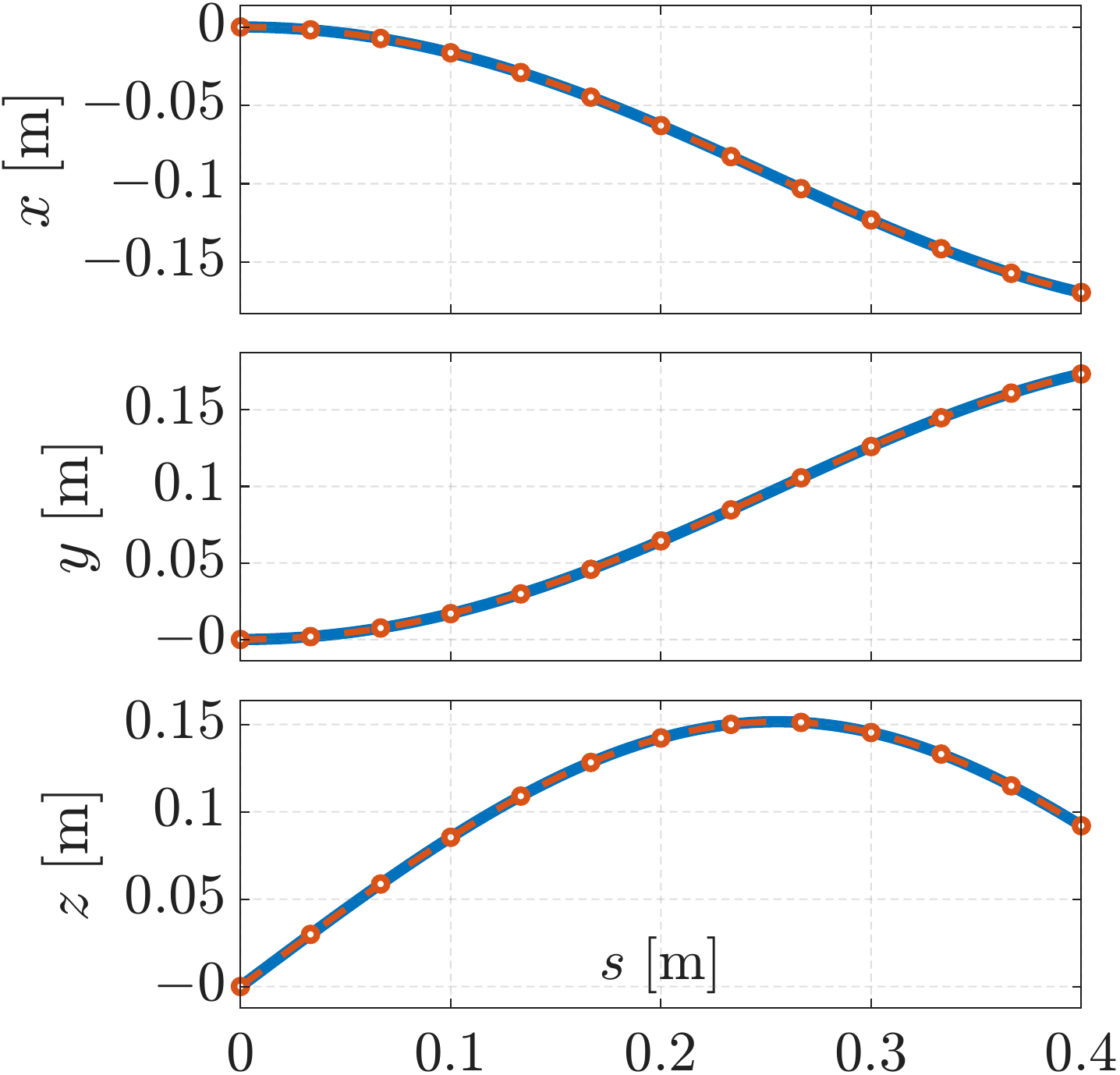}}
\hfill
\subfloat[\(\mathbf F_{\mathrm S}^{(2)}\): \(\boldsymbol{\nu}\)
\label{fig:single-case2-nu}]
{\includegraphics[width=0.245\textwidth]
{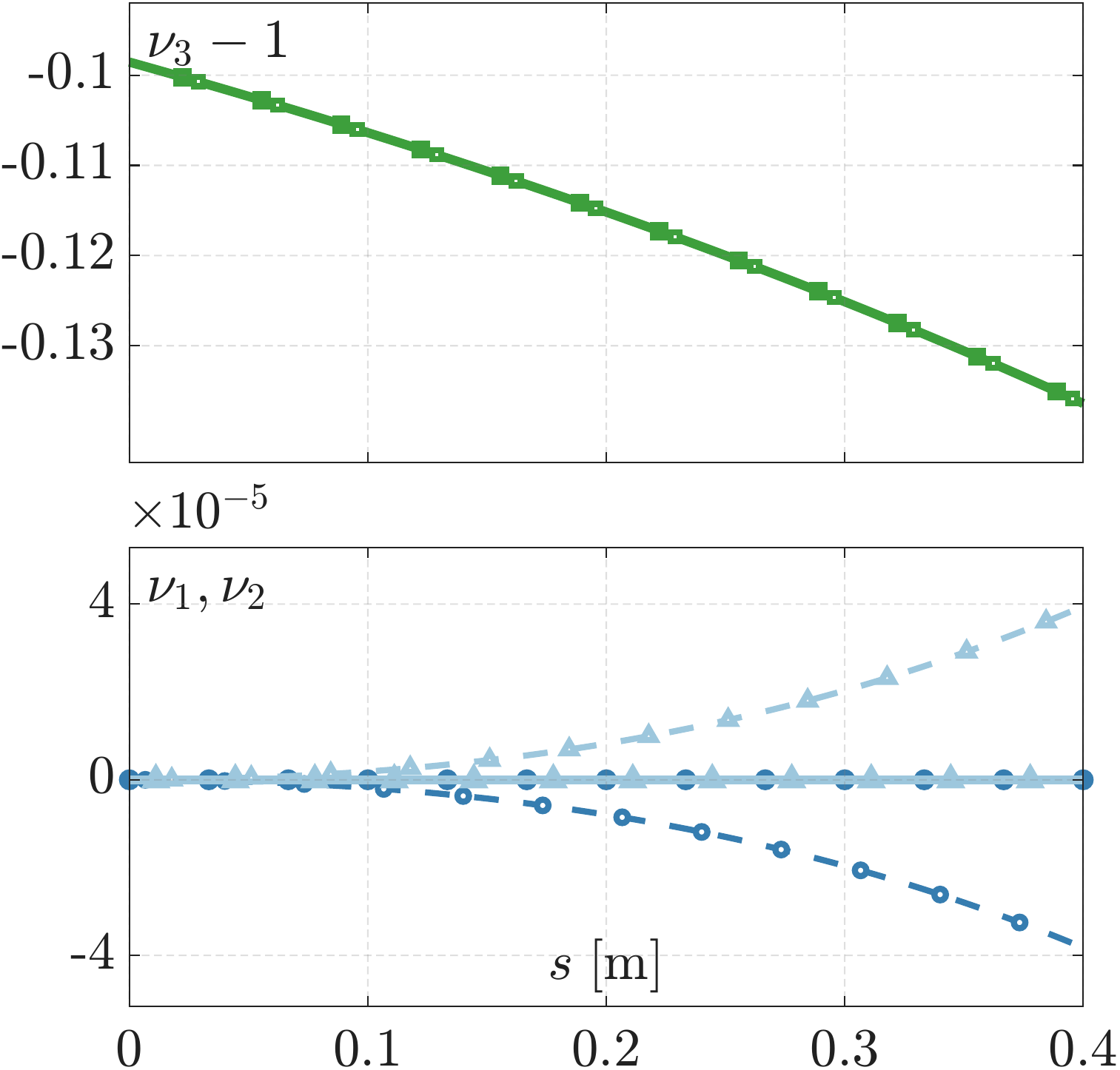}}
\hfill
\subfloat[\(\mathbf F_{\mathrm S}^{(2)}\): \(\kappa_b,\kappa_3\)
\label{fig:single-case2-kappa}]
{\includegraphics[width=0.231\textwidth]
{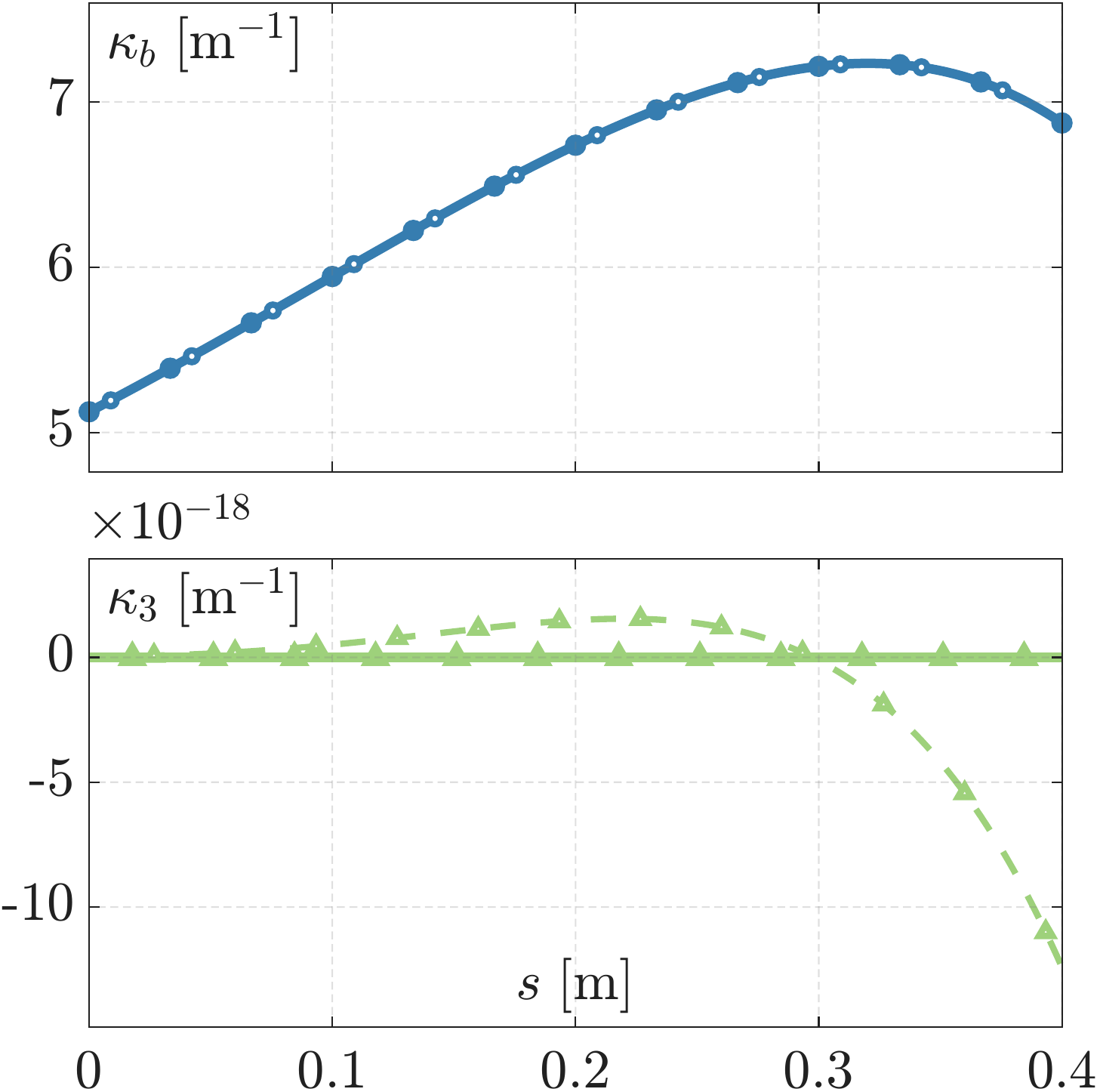}}

\par\vspace{-1.8mm}

\subfloat[\(\mathbf F_{\mathrm S}^{(3)}\): backbone
\label{fig:single-case3-backbone}]
{\includegraphics[width=0.20\textwidth]
{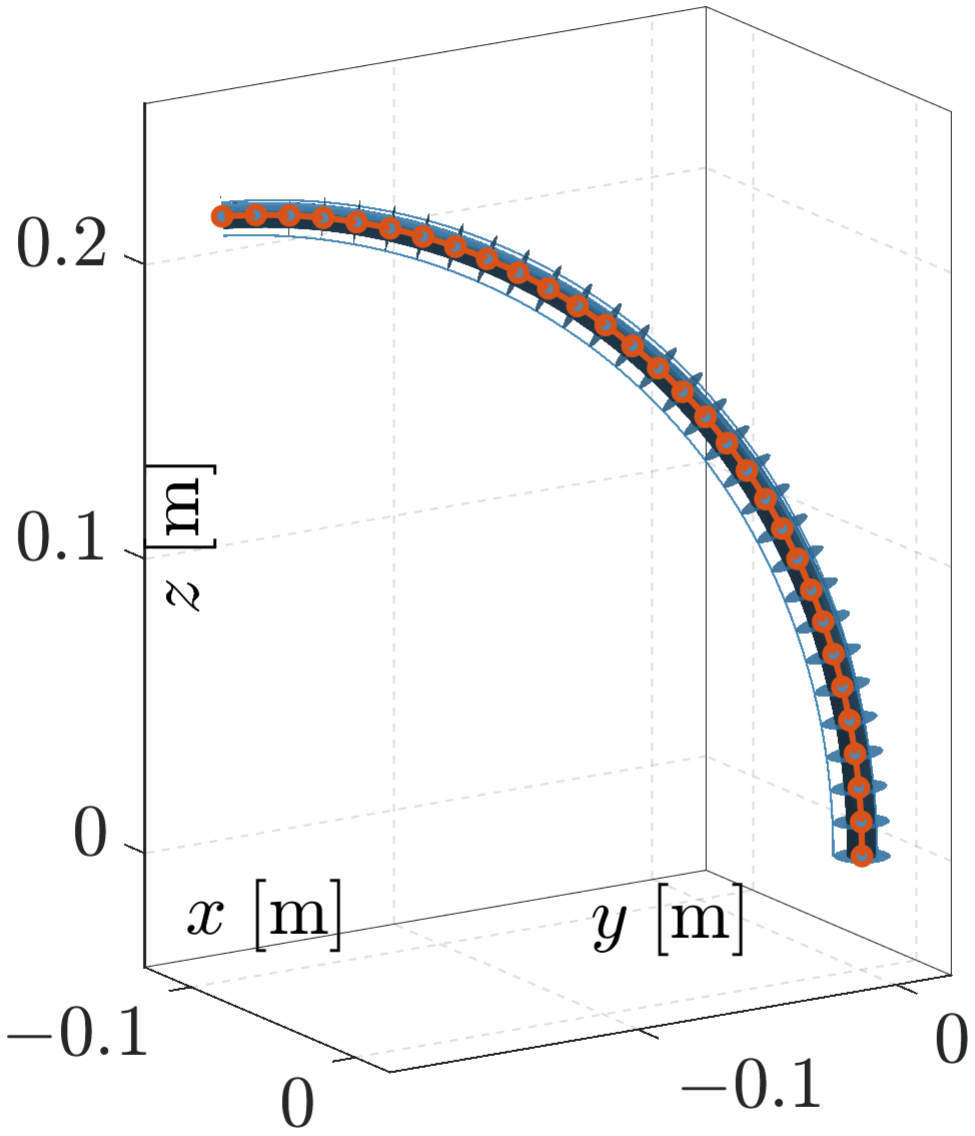}}
\hfill
\subfloat[\(\mathbf F_{\mathrm S}^{(3)}\): \(x,y,z\)
\label{fig:single-case3-cartesian}]
{\includegraphics[width=0.246\textwidth]
{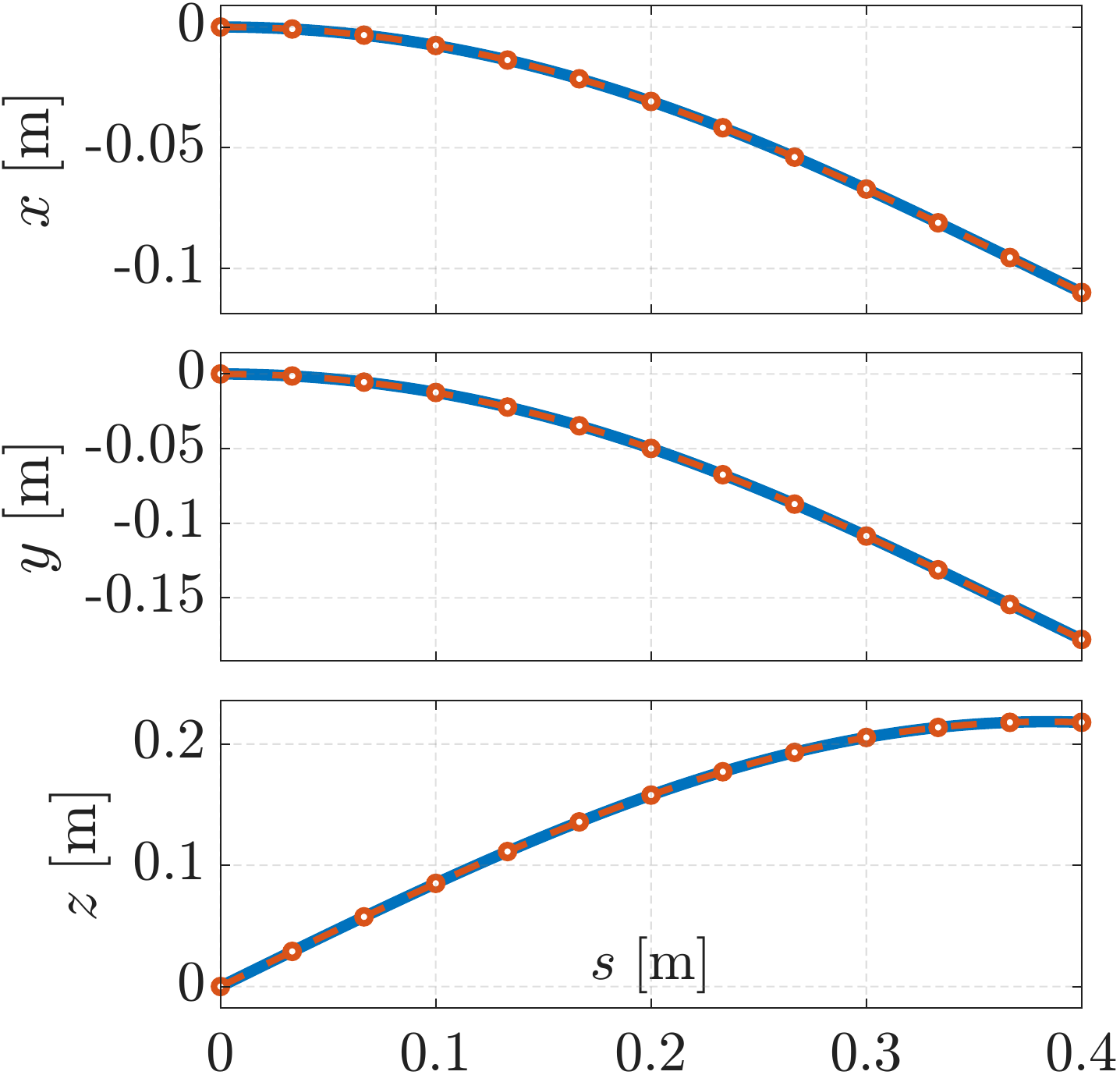}}
\hfill
\subfloat[\(\mathbf F_{\mathrm S}^{(3)}\): \(\boldsymbol{\nu}\)
\label{fig:single-case3-nu}]
{\includegraphics[width=0.247\textwidth]
{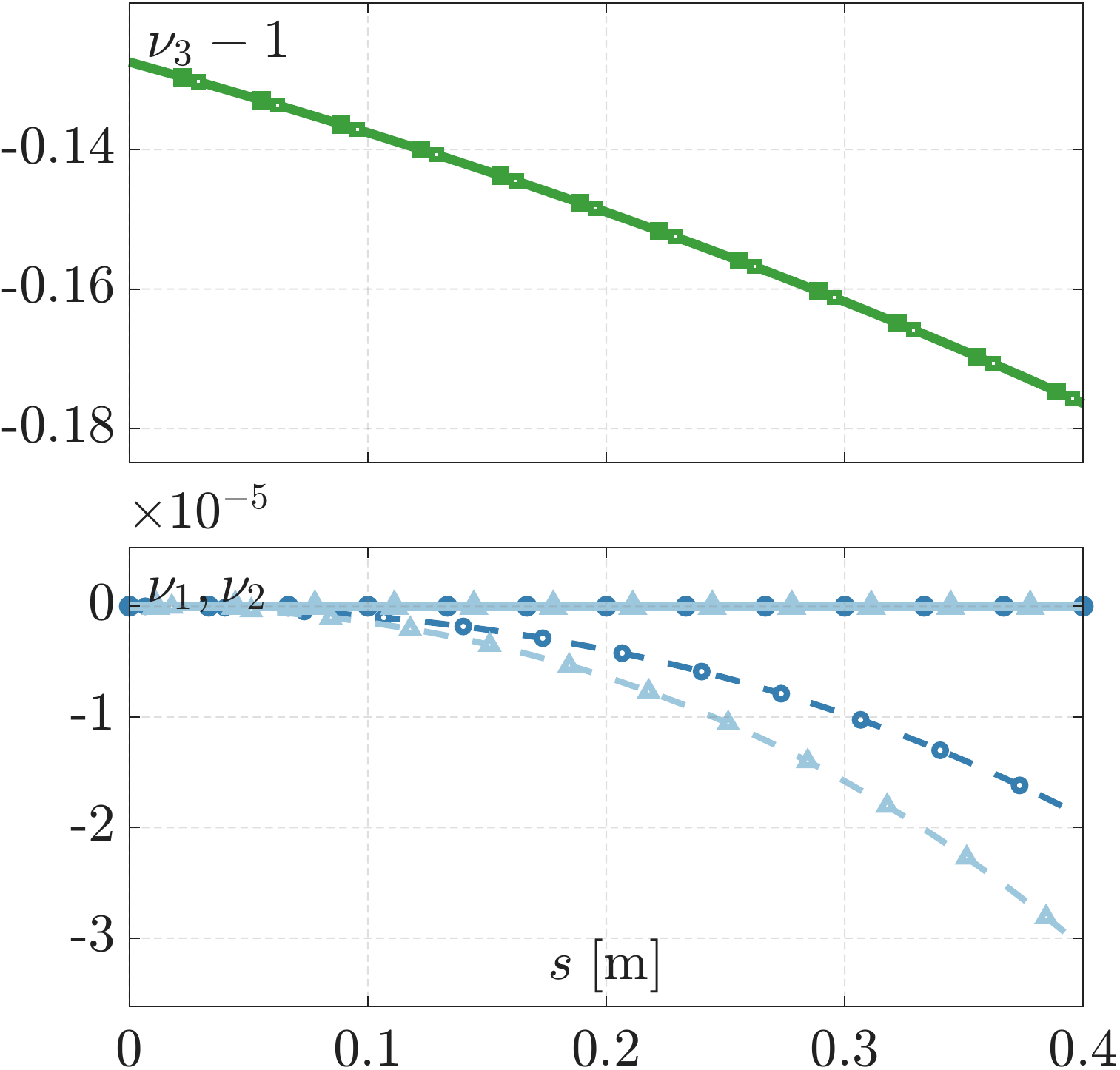}}
\hfill
\subfloat[\(\mathbf F_{\mathrm S}^{(3)}\): \(\kappa_b,\kappa_3\)
\label{fig:single-case3-kappa}]
{\includegraphics[width=0.240\textwidth]
{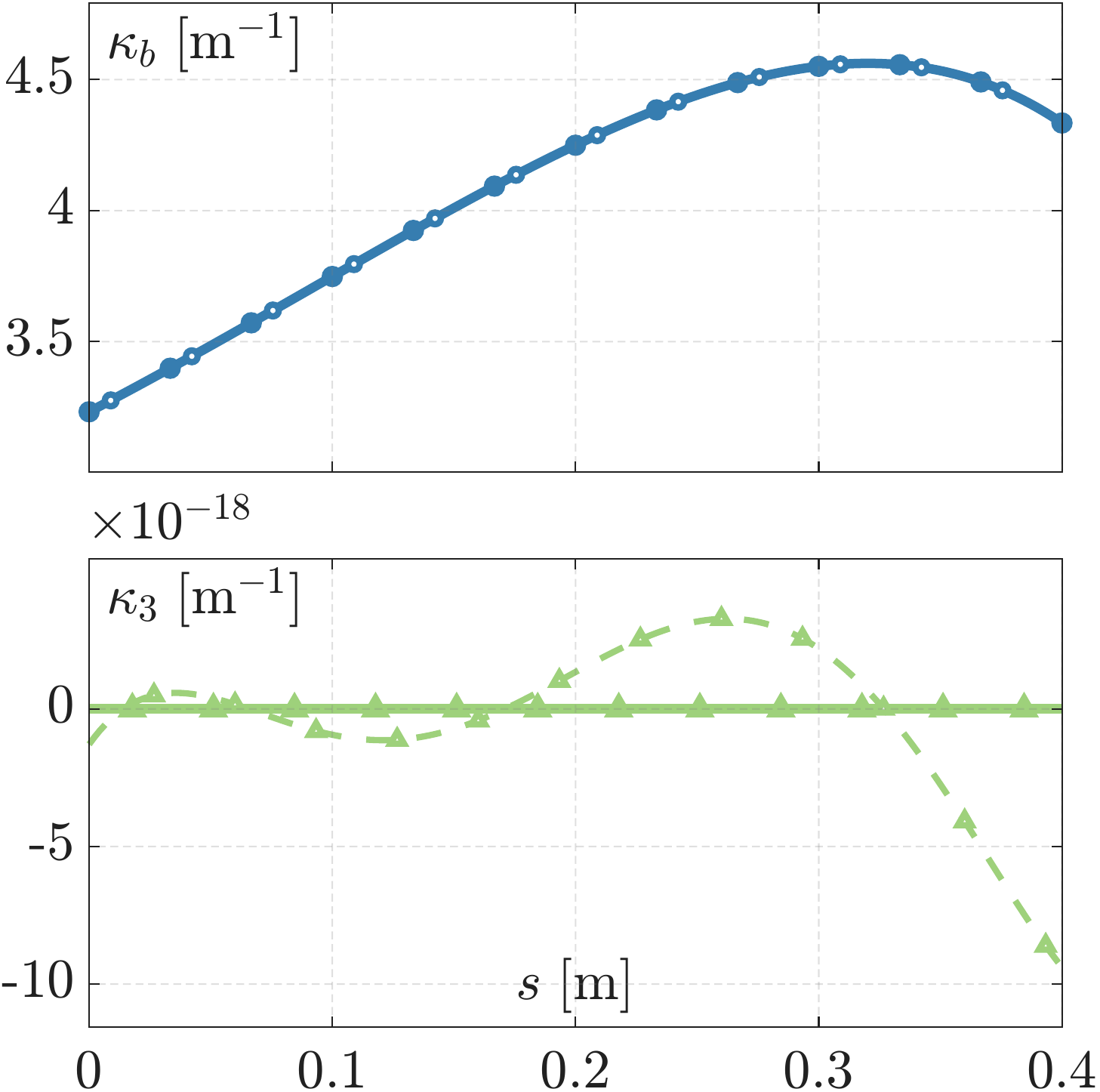}}

\caption{Single-segment comparisons under the three tendon-force cases
in Table~\ref{tab:tendon-force-cases}.}
\label{fig:single-segment-validation}
\end{figure*}
The single-segment case evaluates the combined effects of the
nonuniform cross-sectional geometry and spatially varying
tendon-routing diameter specified in
Table~\ref{tab:simulation-parameters}, with axial deformation retained.
The robot is evaluated under the three tendon-force inputs
\(\mathbf F_{\mathrm S}^{(\ell)}\),
\(\ell=1,2,3\), given in Table~\ref{tab:tendon-force-cases}. Fig.~\ref{fig:single-segment-validation} compares the resulting
Cartesian configurations and distributed strains with GVS. The proposed model captures the changes in bending-plane orientation, bending magnitude, and axial deformation
induced by the different force distributions while
maintaining a continuous Cartesian backbone. The corresponding Cartesian backbone, 
axial-strain and bending-strain profiles also agree closely with those
obtained using GVS. Meanwhile, the small transverse shear strains predicted by
the full-mode GVS reference support neglecting shear in the
slender-backbone formulation introduced in
Sec.~\ref{subsec:generalized-coordinates} for these cases. The material torsional strain is effectively zero throughout the backbone, consistent with the analytical result \(\kappa_3=\eta'=0\) derived from \eqref{eq:twist-equilibrium}.

\subsubsection{Three-Segment Case}
\label{subsubsec:three-segment-case}

\begin{figure*}[!t]
\centering
\captionsetup[subfloat]{font=scriptsize}

\subfloat[\(\mathbf F_{\mathrm M}^{(1)}\): backbone
\label{fig:three-case1-backbone}]
{\includegraphics[width=0.19\textwidth]
{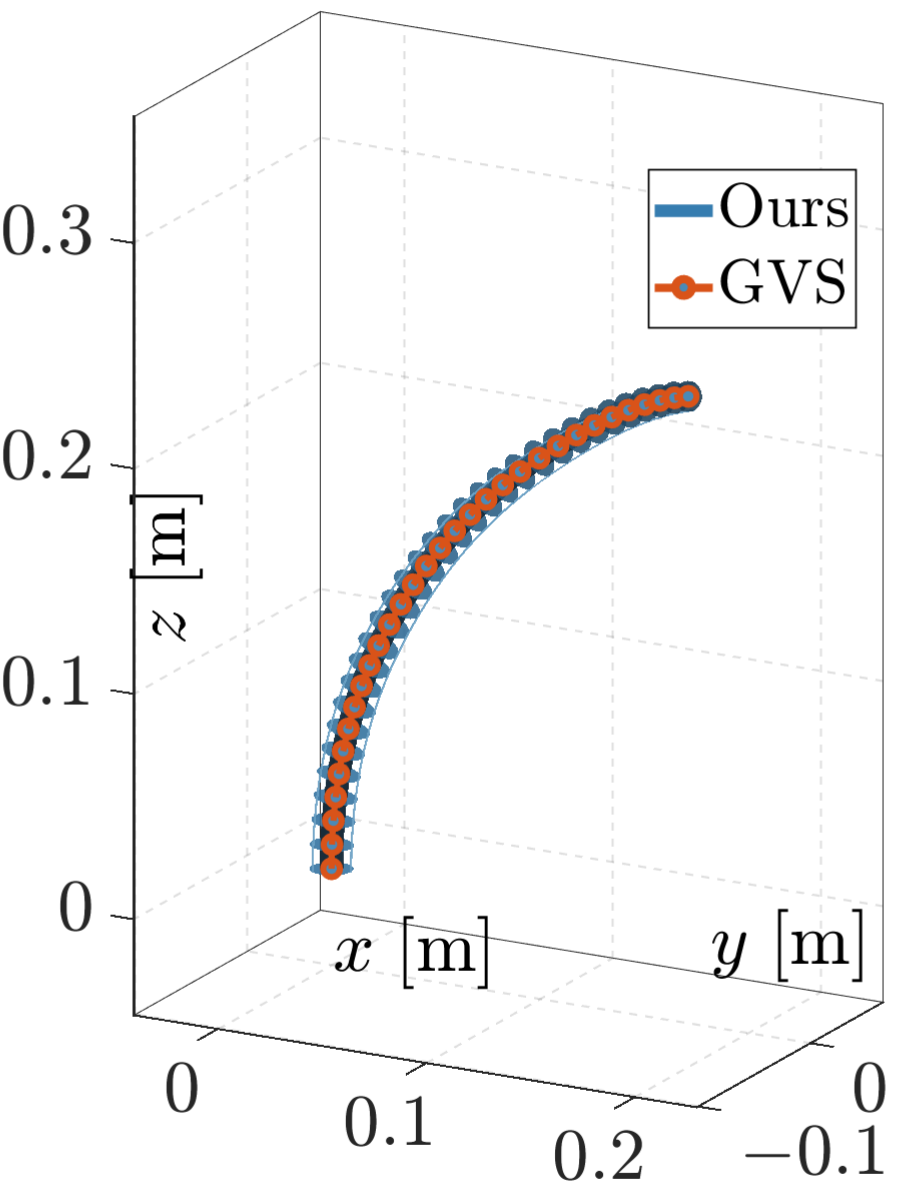}}
\hfill
\subfloat[\(\mathbf F_{\mathrm M}^{(1)}\): \(x,y,z\)
\label{fig:three-case1-cartesian}]
{\includegraphics[width=0.244\textwidth]
{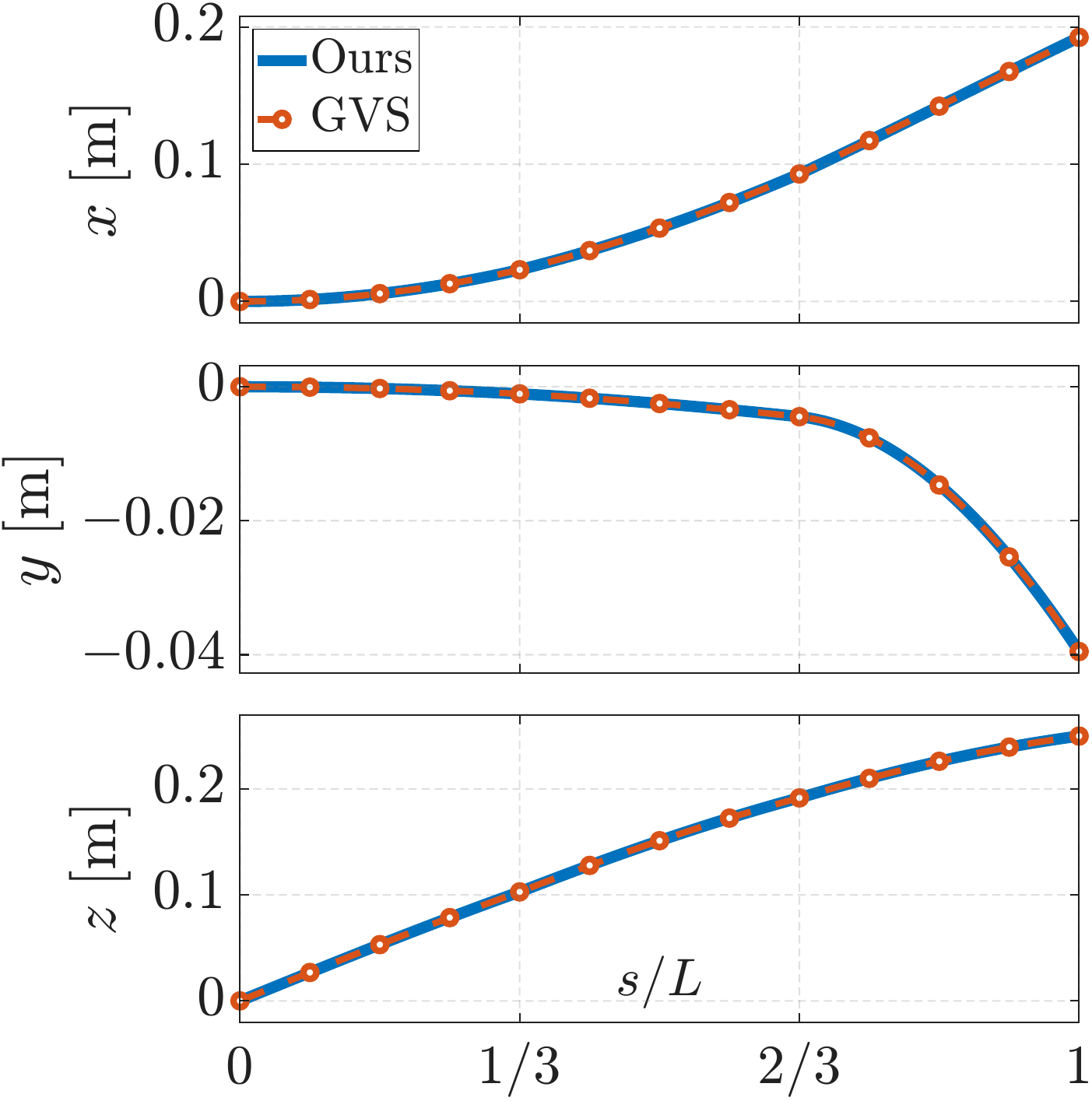}}
\hfill
\subfloat[\(\mathbf F_{\mathrm M}^{(1)}\): \(\boldsymbol{\nu}\)
\label{fig:three-case1-nu}]
{\includegraphics[width=0.245\textwidth]
{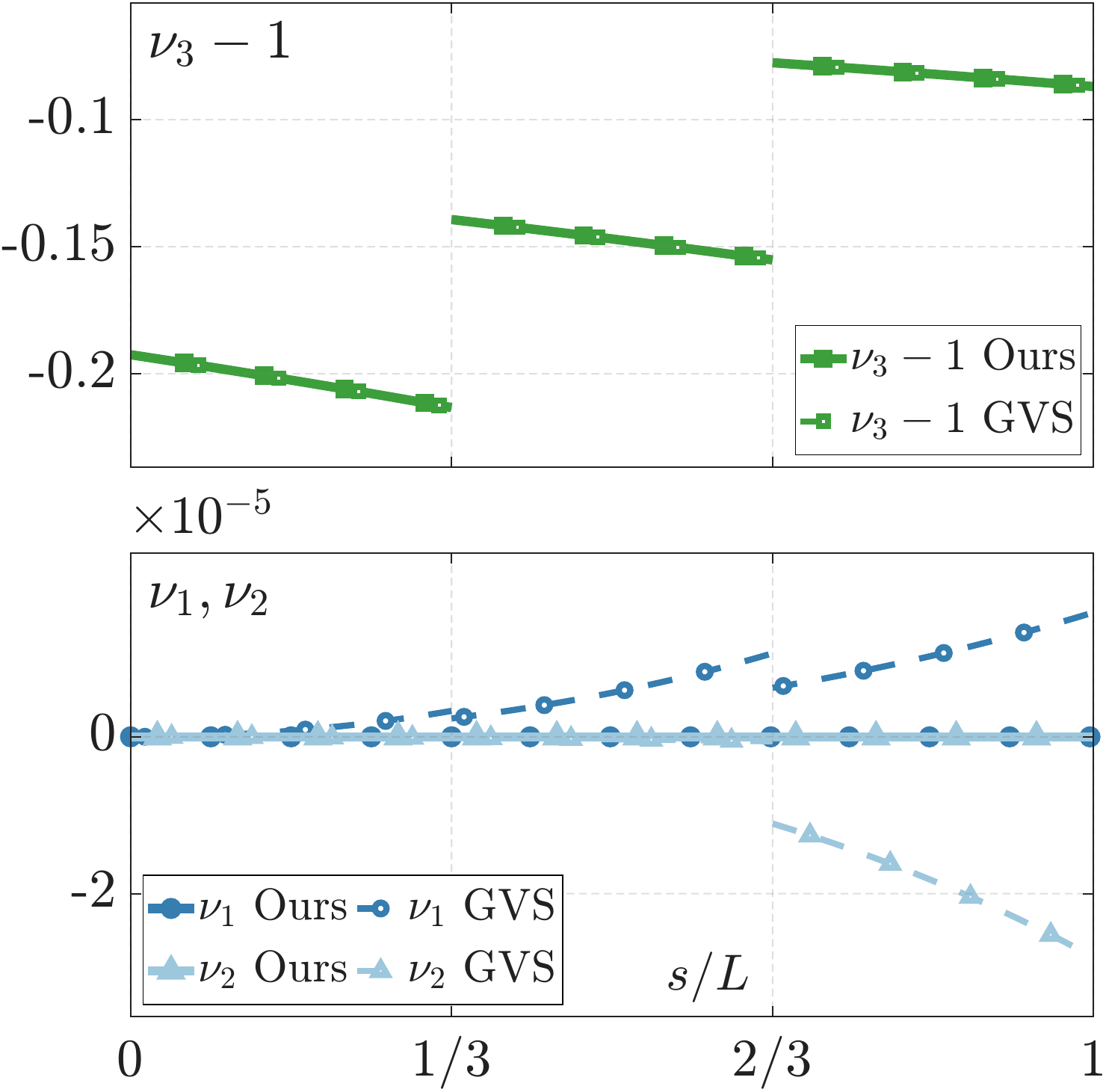}}
\hfill
\subfloat[\(\mathbf F_{\mathrm M}^{(1)}\): \(\kappa_b,\kappa_3\)
\label{fig:three-case1-kappa}]
{\includegraphics[width=0.236\textwidth]
{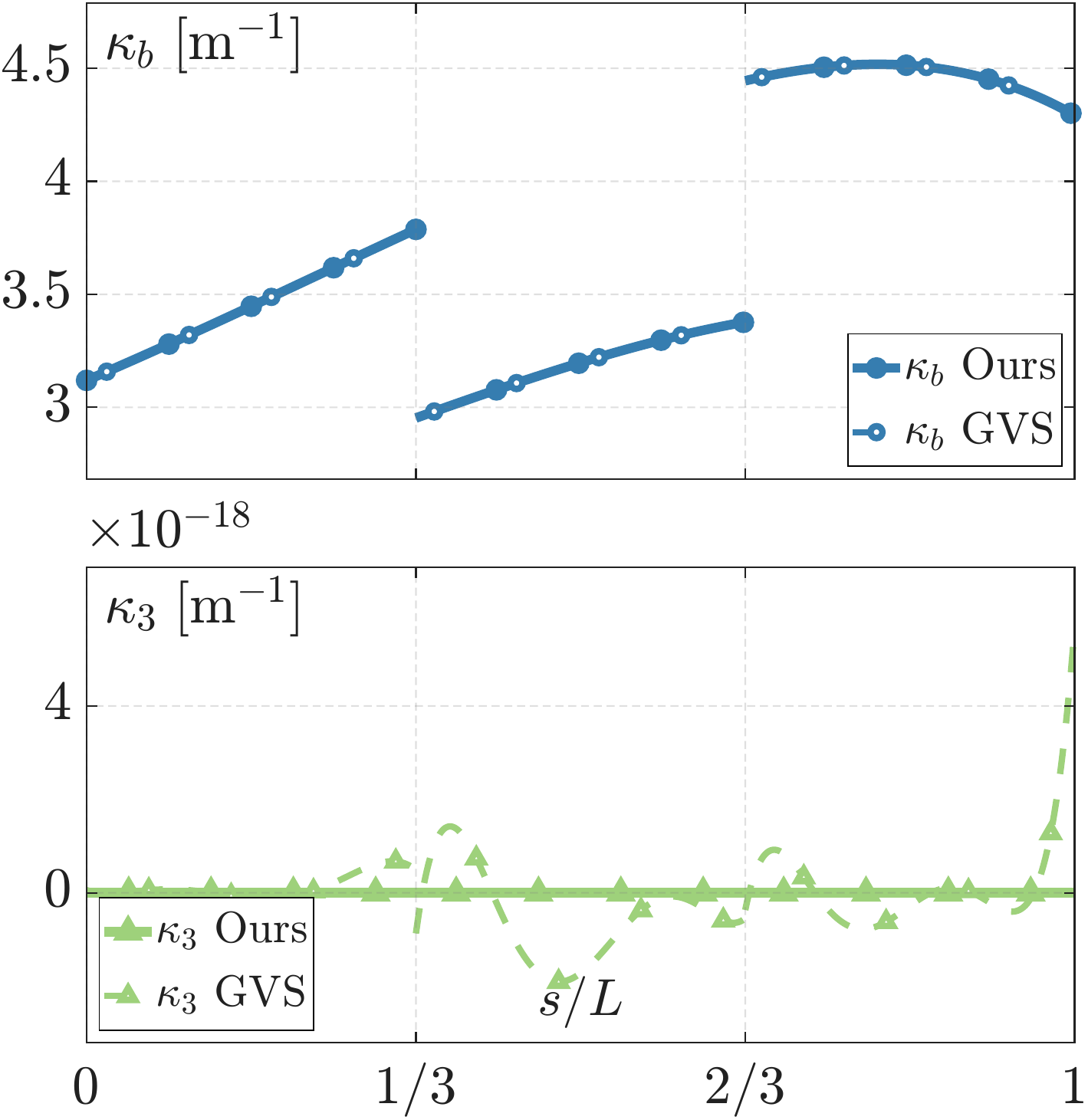}}

\par\vspace{-1.8mm}

\subfloat[\(\mathbf F_{\mathrm M}^{(2)}\): backbone
\label{fig:three-case2-backbone}]
{\includegraphics[width=0.22\textwidth]
{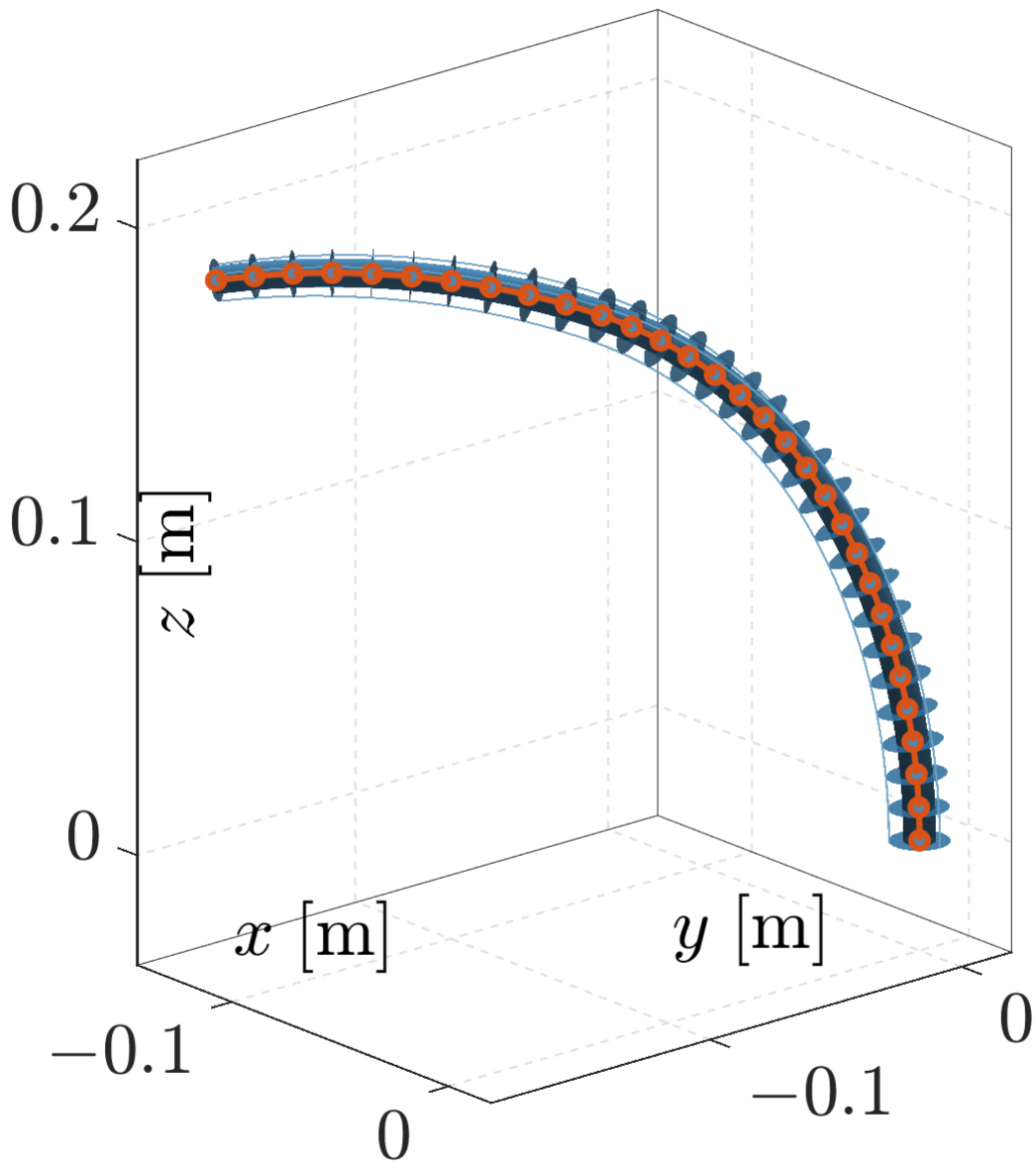}}
\hfill
\subfloat[\(\mathbf F_{\mathrm M}^{(2)}\): \(x,y,z\)
\label{fig:three-case2-cartesian}]
{\includegraphics[width=0.245\textwidth]
{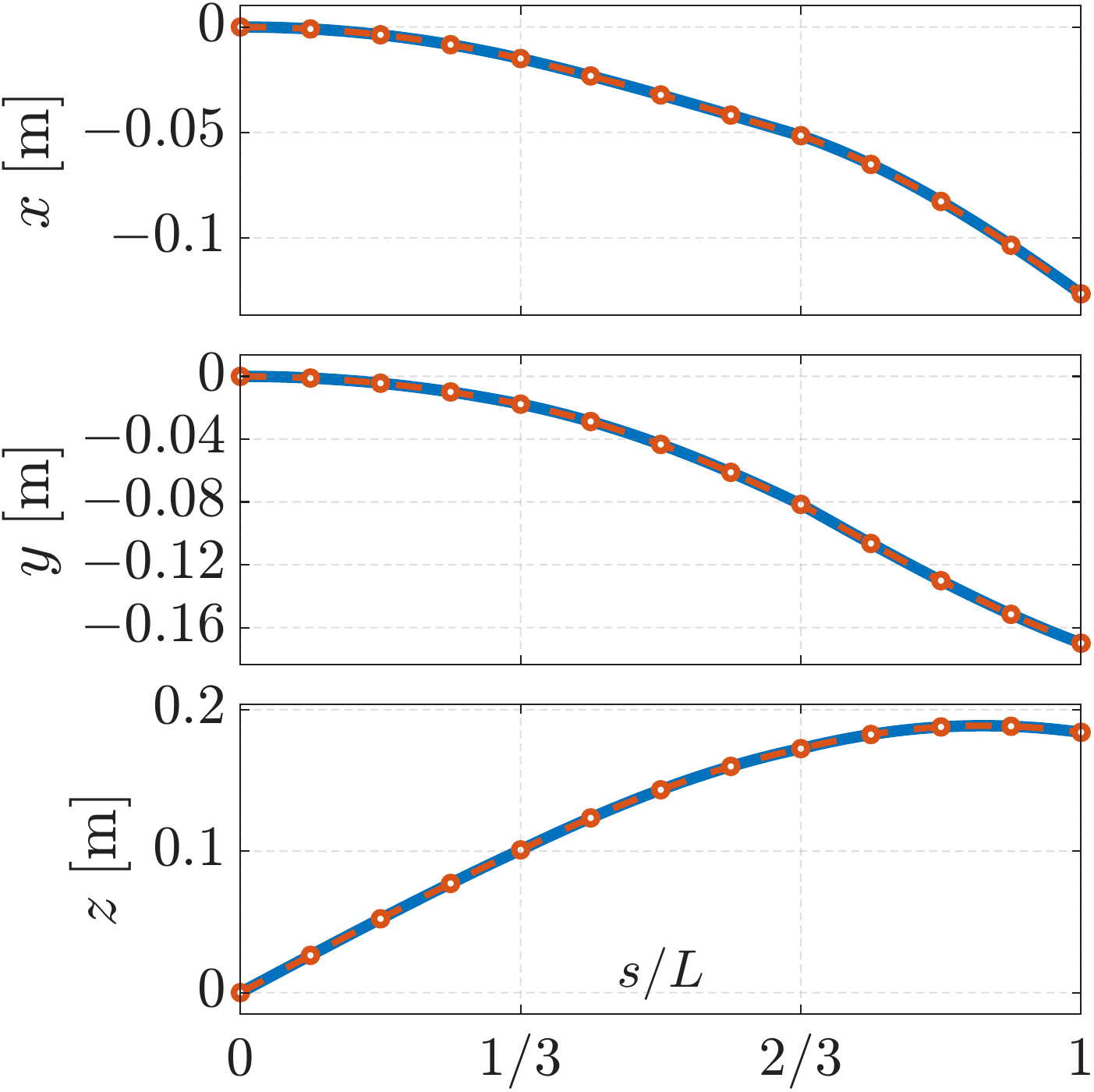}}
\hfill
\subfloat[\(\mathbf F_{\mathrm M}^{(2)}\): \(\boldsymbol{\nu}\)
\label{fig:three-case2-nu}]
{\includegraphics[width=0.245\textwidth]
{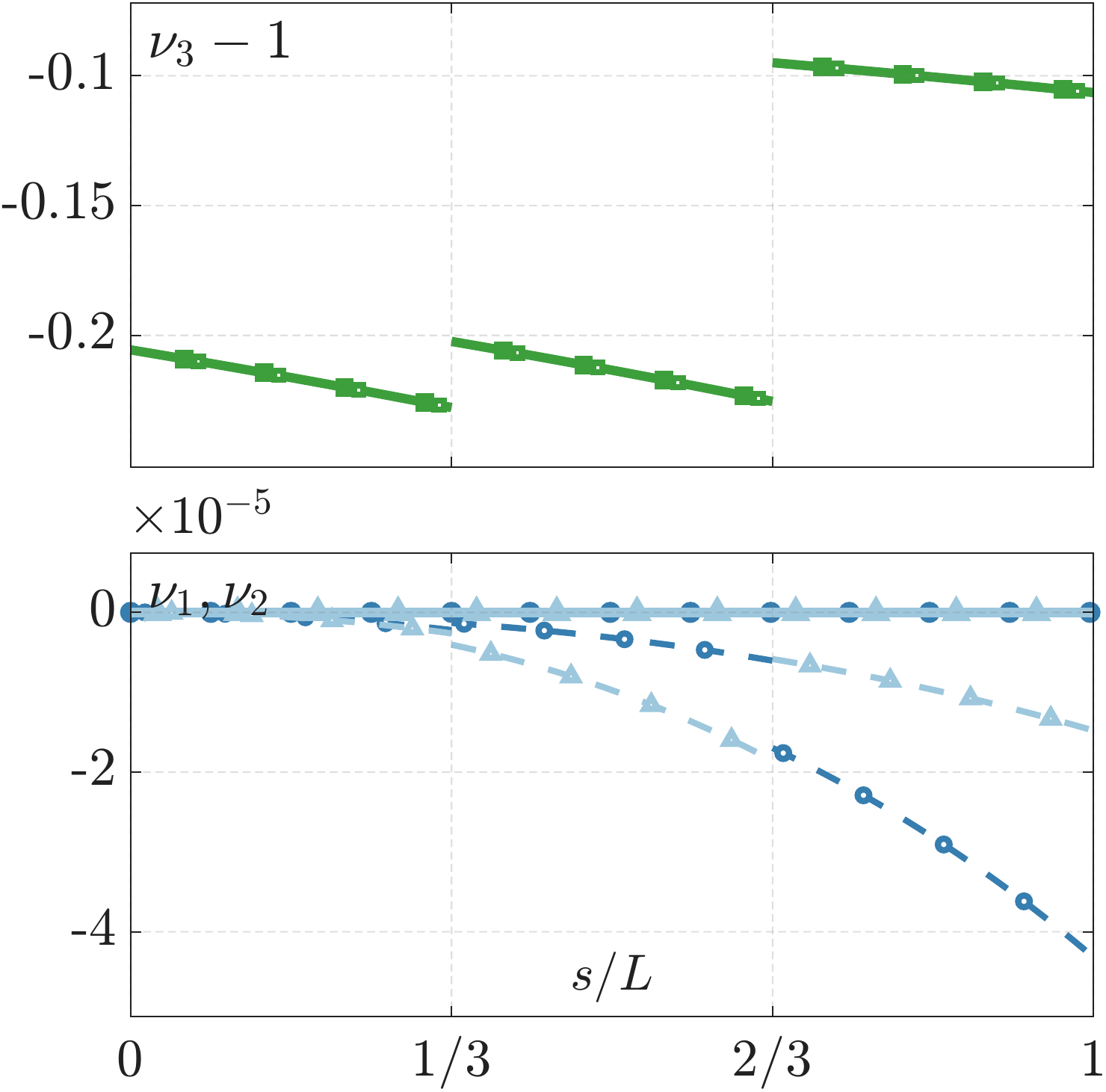}}
\hfill
\subfloat[\(\mathbf F_{\mathrm M}^{(2)}\): \(\kappa_b,\kappa_3\)
\label{fig:three-case2-kappa}]
{\includegraphics[width=0.235\textwidth]
{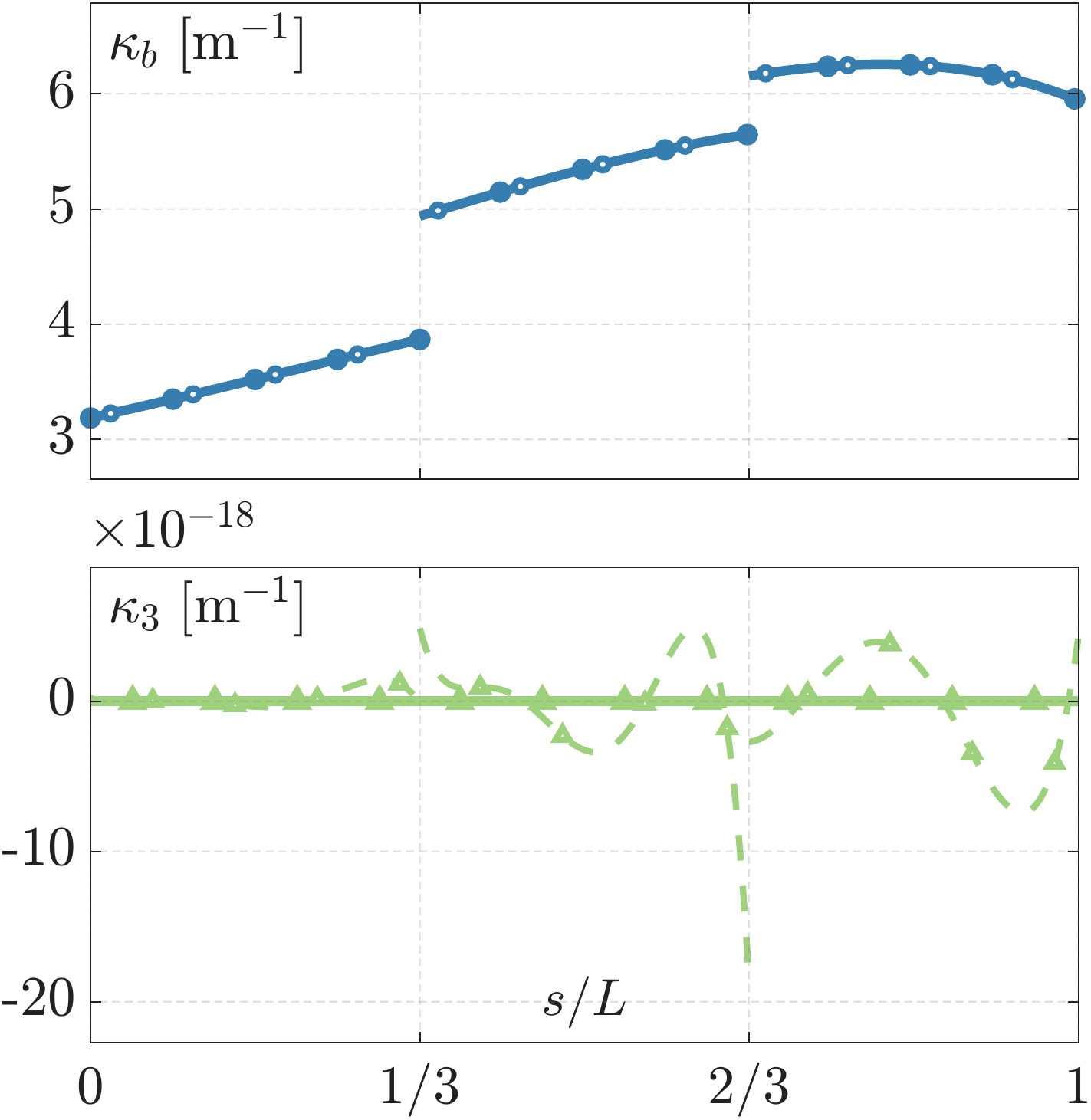}}

\par\vspace{-1.8mm}

\subfloat[\(\mathbf F_{\mathrm M}^{(3)}\): backbone
\label{fig:three-case3-backbone}]
{\includegraphics[width=0.23\textwidth]
{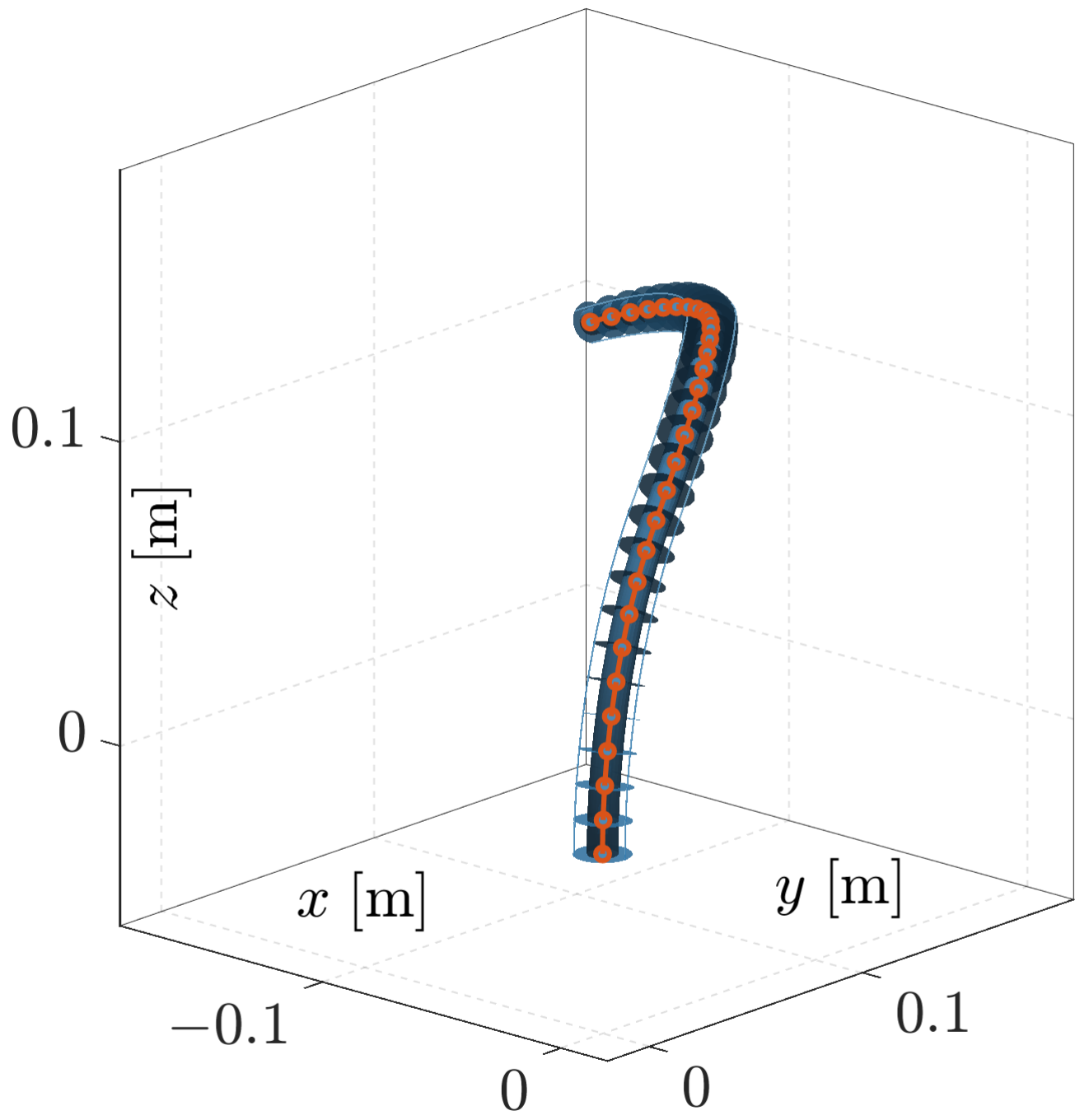}}
\hfill
\subfloat[\(\mathbf F_{\mathrm M}^{(3)}\): \(x,y,z\)
\label{fig:three-case3-cartesian}]
{\includegraphics[width=0.245\textwidth]
{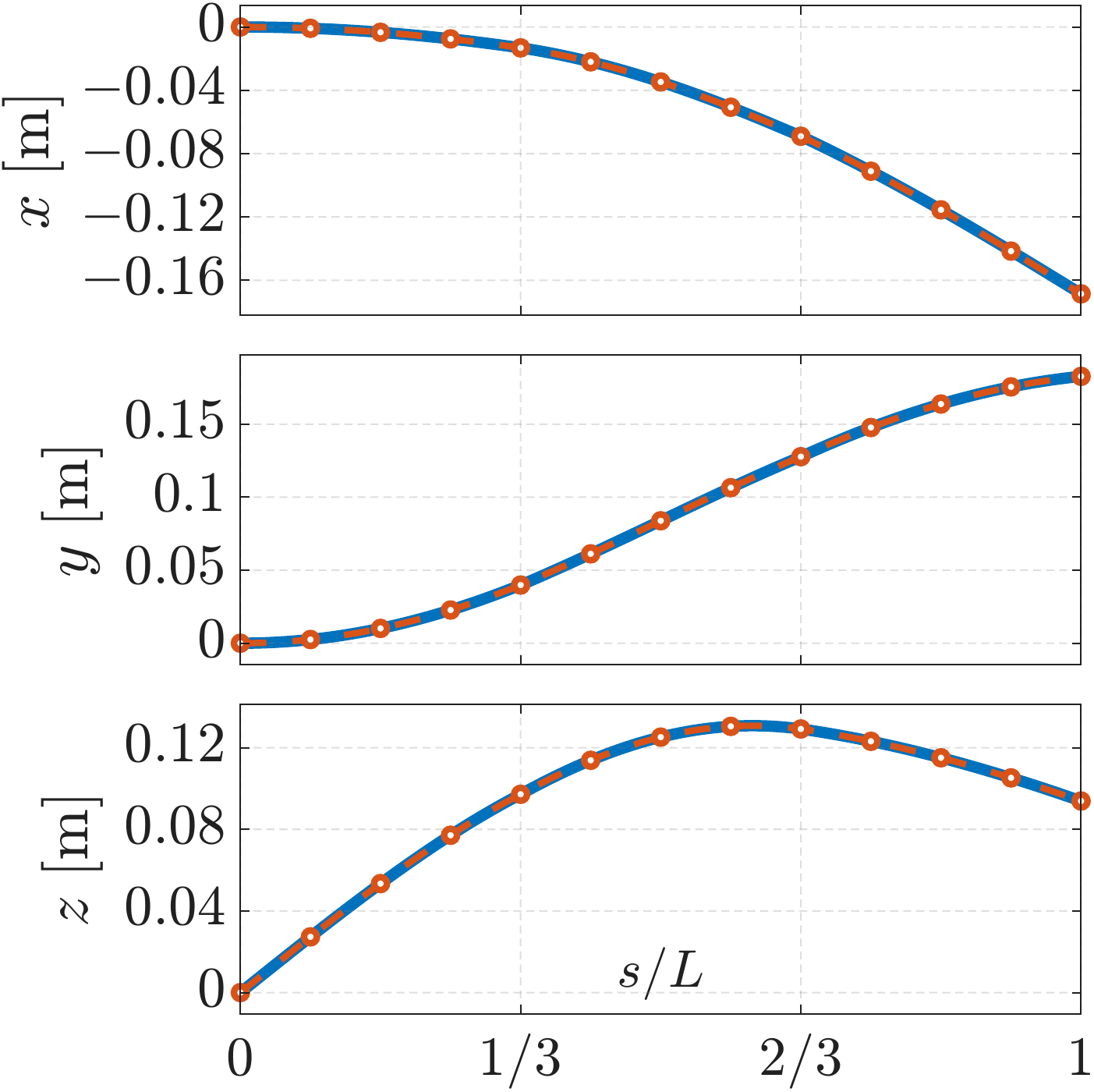}}
\hfill
\subfloat[\(\mathbf F_{\mathrm M}^{(3)}\): \(\boldsymbol{\nu}\)
\label{fig:three-case3-nu}]
{\includegraphics[width=0.245\textwidth]
{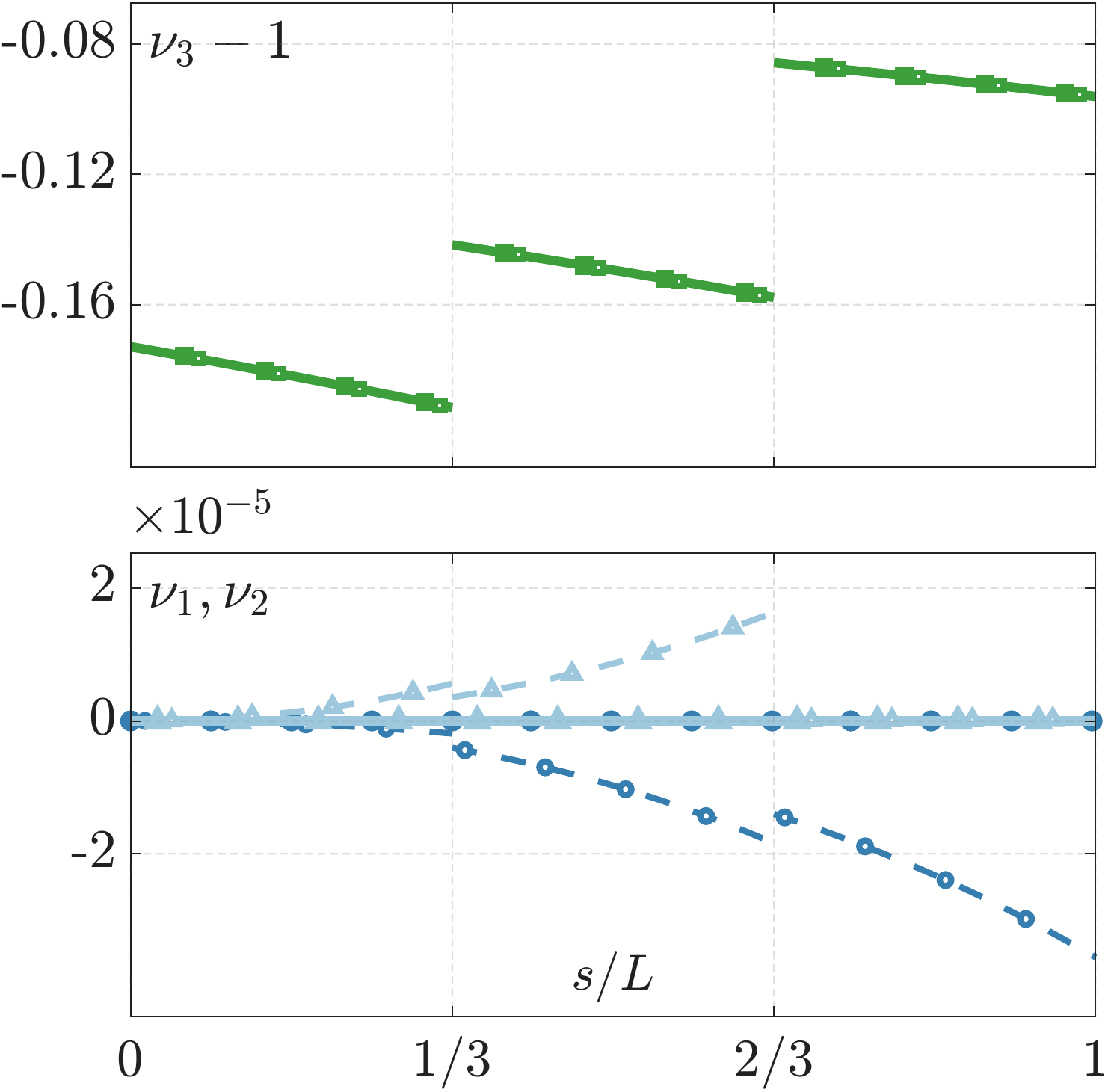}}
\hfill
\subfloat[\(\mathbf F_{\mathrm M}^{(3)}\): \(\kappa_b,\kappa_3\)
\label{fig:three-case3-kappa}]
{\includegraphics[width=0.236\textwidth]
{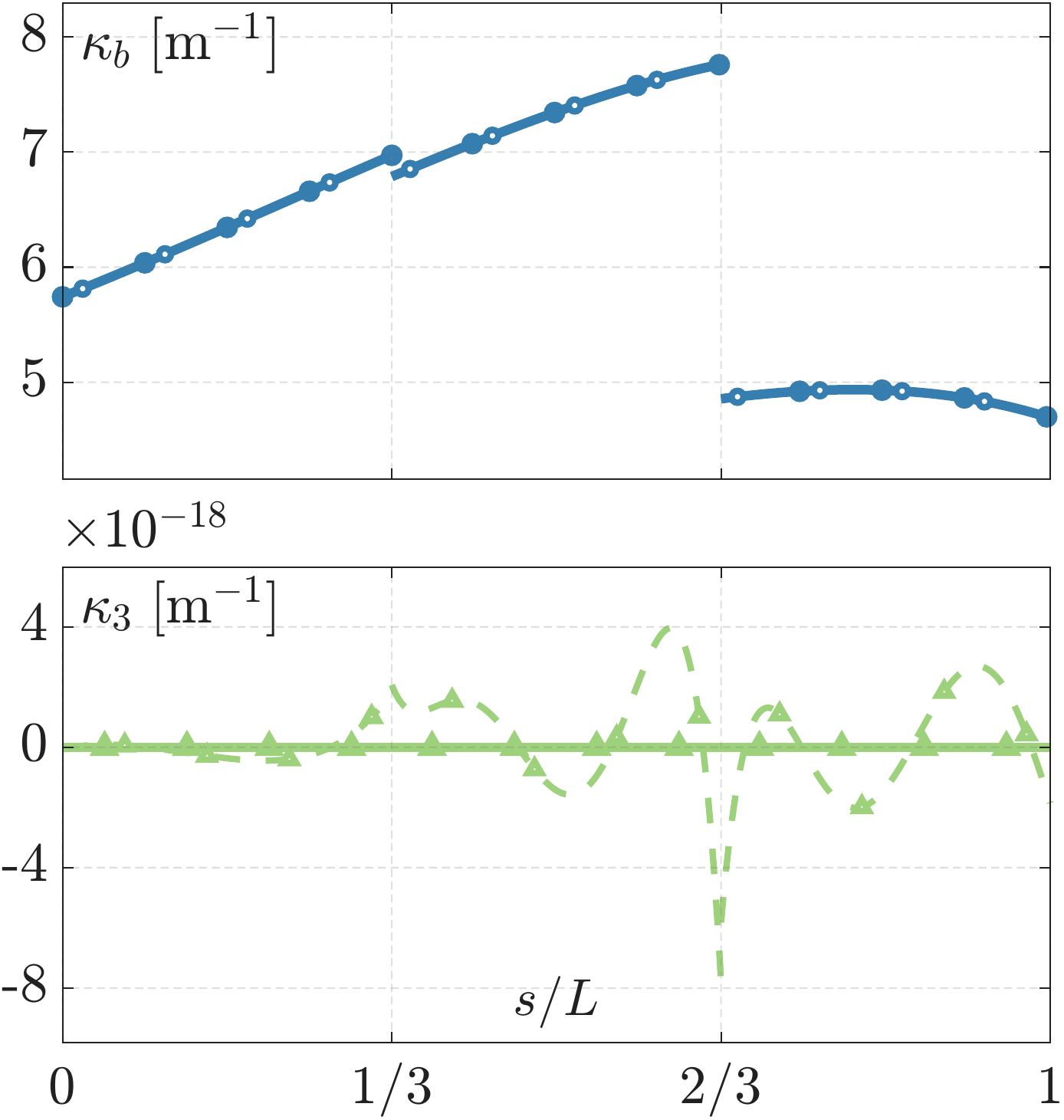}}

\caption{Three-segment comparisons under the tendon-force cases in
Table~\ref{tab:tendon-force-cases}.}
\label{fig:three-segment-validation}
\end{figure*}

The three-segment case evaluates the segmentwise propagation of the
proposed solution when the active tendon set changes across the
interfaces specified in Table~\ref{tab:simulation-parameters}. The
robot is evaluated under the three tendon-force inputs
\(\mathbf F_{\mathrm M}^{(\ell)}\),
\(\ell=1,2,3\), given in Table~\ref{tab:tendon-force-cases}.
Fig.~\ref{fig:three-segment-validation} compares the resulting
Cartesian configurations and distributed strains with GVS. The proposed model also captures the segmentwise axial- and
bending-strain distributions, including their jumps at the segment interfaces, while maintaining continuity of the backbone position and unit tangent. Different segment bending planes combine to produce the overall spatial configuration. The corresponding Cartesian backbone, axial-strain and bending-strain profiles also agree closely with those
obtained using GVS. The transverse shear strains reported by the full-mode GVS
reference remain small within all three segments, supporting
the shear reduction under segment-dependent tendon actuation.
The effectively zero material torsional strain also agrees
with the analytical result \(\kappa_3=\eta'=0\) derived from
\eqref{eq:twist-equilibrium}.

\begin{table}[!t]
\centering
\caption{Workspace analysis and computational performance.}
\label{tab:workspace-performance}
\scriptsize
\setlength{\tabcolsep}{3pt}
\renewcommand{\arraystretch}{1.15}

\begin{tabularx}{\columnwidth}{
    @{}>{\raggedright\arraybackslash}p{0.42\columnwidth}
    >{\centering\arraybackslash}X
    >{\centering\arraybackslash}X@{}
}
\toprule
\textbf{Parameter}
& \textbf{Single-segment continuum robot}
& \textbf{Three-segment continuum robot} \\
\midrule

Tendon-force range (N)
& $[0,0.22]$
& $[0,0.20]$ \\

Tendon-force increment (N)
& $0.01$
& $0.05$ \\

Samples per tendon
& $22$
& $4$ \\

Total force samples
& $\begin{gathered}
N_{\mathrm{W}}=22^3\\
=10648
\end{gathered}$
& $\begin{gathered}
N_{\mathrm{W}}=4^9\\
=262144
\end{gathered}$ \\

Proposed time per point \textbf{($\mu$s)}
& $1.52$
& $2.94$ \\

GVS time per point \textbf{(ms)}
& $2.84$
& $9.85$ \\

Proposed-model speedup over GVS
& $\approx 1864\times$
& $\approx 3348\times$ \\

Maximum values of $e_{\mathrm{tip}}^{(q)}$
& $8.91\times10^{-6}$
& $1.01\times10^{-5}$ \\

\bottomrule
\end{tabularx}

\par\vspace{0.8mm}
\parbox{\columnwidth}{%
\raggedright
\textit{Note:} For each model, the per-point computation time
is the total model-evaluation time divided by $N_{\mathrm{W}}$;
visualization time is excluded.
}
\end{table}

\subsection{Workspace Accuracy and Computational Performance}
\label{subsec:workspace-performance}

The single- and three-segment workspaces are evaluated using
the sampled nonnegative tendon-force combinations summarized
in Table~\ref{tab:workspace-performance}, with
\(N_{\mathrm W}=10{,}648\) and \(262{,}144\), respectively.
Mechanical interference between the backbone and tendons can
further restrict the physically reachable
workspace~\cite{cao2017workspace}.
For each force sample \(\mathbf F^{(q)}\),
\(q=1,\ldots,N_{\mathrm W}\), the length-normalized
tip-position discrepancy relative to GVS is defined as
$$
e_{\mathrm{tip}}^{(q)}
=
\frac{1}{L}
\Bigl\|
\mathbf r_{\mathrm{prop}}
 \bigl(L;\mathbf F^{(q)}\bigr)
-
\mathbf r_{\mathrm{GVS}}
 \bigl(L;\mathbf F^{(q)}\bigr)
\Bigr\|_2.
$$ As reported in Table~\ref{tab:workspace-performance}, the
maximum length-normalized tip-position discrepancies are
\(8.911\times10^{-6}\) and \(1.014\times10^{-5}\) for the
single- and three-segment robots, respectively.
These results indicate close agreement with GVS across the
sampled tendon-force combinations. The proposed model requires average evaluation times of
\(1.52\,\mathrm{\mu s}\) and \(2.94\,\mathrm{\mu s}\) per
configuration, compared with \(2.84\,\mathrm{ms}\) and
\(9.85\,\mathrm{ms}\) for GVS, corresponding to speedups of
approximately \(1864\times\) and \(3348\times\), respectively.
For each model, these times are computed by dividing the
total model-evaluation time by \(N_{\mathrm W}\), with
visualization excluded. The computational advantage follows from the explicit
equilibrium relations and segmentwise Cartesian quadratures,
which avoid an iterative equilibrium solve for each tendon-force
input. The combination of low tip-position discrepancies and
low evaluation costs supports repeated forward-model
evaluations for workspace analysis, motion planning, and
model-based control.
\section{Conclusion}

This work presents a closed-form-in-quadratures
force-to-Cartesian-configuration mapping for spatial
multi-segment TDCRs with nonuniform geometry, variable tendon
routing, and axial deformation.
Across the single- and three-segment cases, the predicted
three-dimensional backbones and distributed strains closely
match the full-mode GVS reference.
The small transverse shear strains support the adopted shear
reduction, while the effectively zero material torsional strain
is consistent with the derived zero-twist equilibrium. Across the sampled tendon-force combinations, the maximum
length-normalized tip-position discrepancies are
\(8.91\times10^{-6}\) and \(1.01\times10^{-5}\) for the
single- and three-segment robots, respectively.
The corresponding average evaluation times are
\(1.52\,\mathrm{\mu s}\) and \(2.94\,\mathrm{\mu s}\),
with speedups of approximately \(1864\times\) and
\(3348\times\) over GVS.
These results support rapid workspace evaluation and repeated
forward-model computations for motion planning and
model-based control.
Future work will extend the formulation to dynamics considering external contacts and forces with experimental validation.

\bibliographystyle{IEEEtran}
\bibliography{reference}
\end{document}